\documentclass[11pt]{article}%
\usepackage{amsmath}
\usepackage{amsfonts}
\usepackage{amssymb}
\usepackage{graphicx}
\usepackage{algorithmic}
\usepackage{algorithm}
\usepackage{hyperref}%
\providecommand{\U}[1]{\protect\rule{.1in}{.1in}}
\providecommand{\U}[1]{\protect\rule{.1in}{.1in}}

\begin{document}

\title{Ensembles of Random SHAPs }
\author{Lev V. Utkin and Andrei V. Konstantinov\\Peter the Great St.Petersburg Polytechnic University\\St.Petersburg, Russia\\e-mail: lev.utkin@gmail.com, andrue.konst@gmail.com}
\date{}
\maketitle

\begin{abstract}
Ensemble-based modifications of the well-known SHapley Additive exPlanations
(SHAP) method for the local explanation of a black-box model are proposed. The modifications
aim to simplify SHAP which is computationally expensive when there is a large
number of features. The main idea behind the proposed modifications is to
approximate SHAP by an ensemble of SHAPs with a smaller number of features.
According to the first modification, called ER-SHAP, several features are
randomly selected many times from the feature set, and Shapley values for the
features are computed by means of \textquotedblleft small\textquotedblright\ SHAPs. 
The explanation results are averaged to get the final Shapley values. 
According to the second modification, called 
ERW-SHAP, several points are generated around the explained instance for
diversity purposes, and results of their explanation are combined with weights
depending on distances between points and the explained instance. The third
modification, called ER-SHAP-RF, uses the random forest for preliminary explanation
of instances and determining a feature probability distribution which is
applied to selection of features in the ensemble-based procedure of ER-SHAP.
Many numerical experiments illustrating the proposed modifications demonstrate
their efficiency and properties for local explanation.


\end{abstract}

\section{Introduction}

Machine learning models and algorithms have shown an increasing importance and
success in many domains. Despite the success, there are obstacles for applying
machine learning algorithms especially in areas of risk, for example, in
medicine, reliability maintenance, autonomous vehicle systems, security
applications. One of the obstacles is that many machine learning models have
sophisticated architectures and, therefore, they are viewed as black boxes. As
a result, models have the limited interpretability, and a user of the
corresponding model cannot understand and explain the predictions and
decisions provided by the model. Another obstacle is that a single testing
instance has to be explained in many cases, i.e., a user needs to understand
only a single prediction, for example, a diagnosis of a patient stated by a
model. In order to overcome these obstacles, additional interpretable models
should be developed that could help to answer the question, which features of
an analyzed instance lead to the black-box survival model prediction. In other
words, these models should select the most important features which impact on
the black-box model prediction. It should be noted that some models, including
linear regression, logistic regression, decision trees are intrinsically
explainable due to their peculiarities. At the same time, most machine
learning models, especially, deep learning models are black boxes and cannot
be directly explained. Explanation of these models and their predictions
motivated developing a lot of methods and models which try to explain
predictions of the deep classification and regression algorithms. There are
several detailed survey papers providing a deep dive into variety of
interpretation methods and models
\cite{Belle-Papantonis-2020,Guidotti-2019,Liang-etal-2021,Marcinkevics-Vogt-20,Molnar-2019,Xie-Ras-etal-2020,Zablocki-etal-21,Zhang-Tino-etal-2020}%
, which show an increasing importance of the interpretation methods and a
growing interest to them.

Interpretation of the black-model local prediction aims to select features
which significantly impact on this prediction, i.e., by using the
interpretation model, we try to determine \textit{which features of an
analyzed instance lead the obtained black-box model prediction}. There are two
groups of the interpretation methods. The first one consists of the so-called
local methods. They try to interpret a black-box model locally around a test
instance. The second group contains global methods which derive interpretations
on the whole dataset or its part. The present paper focuses on the first group
of the local interpretation methods though the proposed approach can be simply
extended to the global interpretation.

Two very popular post-hoc approaches to interpretation can be selected among
many others. The first one is LIME (Local Interpretable Model-Agnostic
Explanation) \cite{Ribeiro-etal-2016}, which is based on building an
approximating linear model around the instance to be explained. This follows
from the intuition that the explanation may be derived locally from many
instances generated in the neighborhood of the explained instance with weights
defined by their distances from the explained instance. Coefficients of the
linear model are interpreted as the feature's importance. The linear
regression for solving the regression problem or logistic regression for
solving the classification problem allow us to construct the corresponding
linear models. LIME\ has many advantages. It successfully interprets models
dealing with tabular data, text, and images. However, there are some
shortcomings of LIME. The first one is that LIME is not robust. This means
that it may provide very different explanations for two nearby data points.
The definition of neighborhoods is also very vague. Moreover, LIME may provide
incorrect explanation when there is a small difference between training and
testing data. LIME is also sensitive to parameters of the explanation model,
for example, to weights of generated instances, to the number of the generated
instances, etc.

The second approach consists of the well-known method SHAP (SHapley Additive
exPlanations) \cite{Lundberg-Lee-2017,Strumbel-Kononenko-2010} and its
modifications. The method is inspired by game-theoretic Shapley values
\cite{Shapley-1953} which can be interpreted as average expected marginal
contributions over all possible subsets (coalitions) of features to the
black-box model prediction. SHAP has many advantages, for example, it can be
used for local and global explanations in contrast to LIME, but there are also
two important shortcomings. The first one is a question how to add or remove
features in order to implement their subsets as inputs for the black-box
model. There are many approaches to removing features, exhaustively described
by Covert et al. \cite{Covert-etal-20}, but SHAP may be too sensitive to each
of them, and there is no strong justifications of their use. Nevertheless,
SHAP can be regarded as the most promising and efficient explanation method.

The second shortcoming is that SHAP is computationally expensive when there is
a large number of features due to considering all possible coalitions whose
number is $2^{m}$, where $m$ is the number of features. Therefore, the
computational time grows exponentially. Several simplifications and
approximations have been proposed in order to overcome this difficulty. Some
of them are presented by Strumbelj and Kononenko
\cite{Strumbel-Kononenko-2010,Strumbelj-Kononenko-11,Strumbel-Kononenko-14}.
One of the simplifications is based on using ordered permutations of the
feature indices and probability distributions of features
\cite{Strumbelj-Kononenko-11}. Another approximation is the quasi-random and
adaptive sampling which includes two improvements \cite{Strumbel-Kononenko-14}%
. The first one is based on exploiting Monte Carlo integration. The second
improvement is based on the optimal number of perturbations of every feature
in accordance with its variance to minimize the overall approximation error.
Strumbelj and Kononenko \cite{Strumbel-Kononenko-14} also proposed to average
local contributions of values of each feature across all instances. Another
interesting approach to simplify SHAP is the Kernel SHAP
\cite{Lundberg-Lee-2017} which can be regarded as a computationally efficient
approximation to Shapley values in higher dimensions. In order to relax the
assumption of the feature independence accepted in Kernel SHAP, Aas et al.
\cite{Aas-etal-2019} extended the Kernel SHAP method to handle dependent
features. Ancona et al. \cite{Ancona-etal-2019} proposed the polynomial-time
approximation of Shapley values called the Deep Approximate Shapley
Propagation (DASP) method.

In spite of many approaches to simplify SHAP, it is difficult to expect a
significant simplification from the above modifications of SHAP. Therefore, a
new approach is proposed for simplifying the SHAP method and for reducing
computational expenses for calculating the Shapley values. A key idea behind
the proposed approach is to apply a modification of the random subspace method
\cite{Ho-98} and to consider an ensemble of random SHAPs. The approach is very
similar to the random forests when an ensemble of randomly built decision
trees is used to get some average classification or regression measures.
Random SHAPs are constructed by random selection of $t$ features with indices
$J_{k}=(i_{1},...,i_{t})$ from the instance for explanation and the obtained
subset of the instance features is analyzed by SHAP as a separate instance.
Repeating this procedure $N$ times, we get a set $\mathcal{S}=\{S_{1}%
,...,S_{N}\}$ of the Shapley values corresponding to the input subsets of
features, where the $k$-th subset is $S_{k}=\{\phi_{i}:i\in J_{k}\}$. By
applying some combination rule for combining subsets $S_{k}$ from
$\mathcal{S}$, we obtain the final Shapley values.

The above general approach considering an ensemble of random SHAPs has several
extensions which form the corresponding methods and algorithms. First of all,
we can generate points around the analyzed instance and construct $S_{k}$ for
the $k$-th generated point. In this case, every point is assigned by a weight
depending on the distance from the analyzed point. As a result, we can combine
subsets $S_{k}$ of the Shapley values with weights which are defined as a
function of the distance from the analyzed point.

Another extension or modification is to select features in accordance with a
probability distribution to get instances consisting of features with indices
from the set $J_{k}$. Let us define the discrete probability distribution over
the set of all indices. It can be produced, for example, by using the random
forest \cite{Breiman-2001} which plays a role of a feature selection model. At
that, the random forest is constructed by using a set of points (instances)
locally generated around the explained point. Every decision tree is built by
using a single point from the set of generated points.

In sum, the contribution of the paper can be formulated as follows:

\begin{enumerate}
\item A new approach to implementing an ensemble-based SHAP with random
subsets of features of the explained instance is proposed.

\item Several combination schemes are studied for aggregating subsets of
important features obtained by using random SHAPs.

\item The approach is extended by generating random points in the local area
around a test instance and computing subsets of important features separately
for every point. A some kind of diversity is implemented with this extension.

\item Another extension is to use a probability distribution for the random
selection of features defined by means of the random forest constructed by
using the generated points in the local area around a test instance. The
preliminary feature selection can be regarded as a pre-training procedure.
\end{enumerate}

A lot of numerical experiments with an algorithm implementing the proposed
method on synthetic and real datasets demonstrate its efficiency and
properties for local and global interpretation.

The paper is organized as follows. Related work is in Section 2. The Shapley
values and the SHAP method as a powerful tool for local and global
explanations are introduced in Section 3. A detailed description of the
proposed modifications of SHAP, including ER-SHAP, ERW-SHAP, ER-SHAP-RF is
provided in Section 4. Numerical experiments with synthetic data and real data
using the local interpretation by means of the proposed models and their
comparison with the standard SHAP method are given in Section 5. Concluding
remarks can be found in Section 6.

\section{Related work}

\textbf{Local interpretation methods. }An increasing importance of machine
learning models and algorithms leads to development of new explanation methods
taking into account various peculiarities of applied problems. As a result,
many models of the local interpretation have been proposed. Success and
simplicity of the LIME interpretation method resulted in development of
several its modifications, for example, ALIME
\cite{Shankaranarayana-Runje-2019}, Anchor LIME \cite{Ribeiro-etal-2018},
LIME-Aleph \cite{Rabold-etal-2019}, GraphLIME \cite{Huang-Yamada-etal-2020},
SurvLIME \cite{Kovalev-Utkin-Kasimov-20a}, etc. A comprehensive analysis of
LIME, including the study of its applicability to different data types, for
example, text and image data, is provided by Garreau and Luxburg
\cite{Garreau-Luxburg-2020}. The same analysis for tabular data is proposed by
the same authors \cite{Garreau-Luxburg-2020a}. An image version of LIME with
its deep theoretical study is presented by Garreau and Mardaoui
\cite{Garreau-Mardaoui-21}. An interesting information-theoretic justification
of interpretation methods on the basis of the concept of the explainable
empirical risk minimization is proposed by Jung \cite{Jung-20}.

In order to relax the linearity condition for the local interpretation models
like LIME and to adequately approximate a black-box model, several
interpretation methods based on using Generalized Additive Models (GAMs)
\cite{Hastie-Tibshirani-1990} were proposed
\cite{Chang-Tan-etal-2020,Lou-etal-12,Nori-etal-19,Zhang-Tan-Koch-etal-19}.
Another interesting class of models based on using a linear combination of
neural networks such that a single feature is fed to each network was proposed
by Agarwal et al. \cite{Agarwal-etal-20}. The impact of every feature on the
prediction in these models is determined by its corresponding shape function
obtained by each neural network. Following ideas behind these interpretation
models, Konstantinov and Utkin \cite{Konstantinov-Utkin-20b} proposed a
similar model. In contrast to the method proposed by Agarwal et al.
\cite{Agarwal-etal-20}, an ensemble of gradient boosting machines is used in
\cite{Konstantinov-Utkin-20b} instead of neural networks in order to simplify
the explanation model training process.

Another explanation method is SHAP
\cite{Lundberg-Lee-2017,Strumbel-Kononenko-2010}, which takes a game-theoretic
approach for optimizing a regression loss function based on Shapley values.
General questions of the computational efficiency of SHAP were investigated by
Van den Broeck et al. \cite{Broeck-etal-21}. Bowen and Ungar
\cite{Bowen-Ungar-20} proposed the generalized SHAP method (G-SHAP) which
allows us to compute the feature importance of any function of a model's
output. Rozemberczki and Sarkar \cite{Rozemberczki-Sarkar-21} presented an
approach to applying SHAP to ensemble models. The problem of explaining the
predictions of graph neural networks by using SHAP was considered by Yuan et
al. \cite{Yuan-Yu-20}. Frye et al. \cite{Frye-etal-2020} introduced the
so-called off- and on-manifold Shapley values for high-dimensional multi-type
data. Application of SHAP to explanation of recurrent neural networks was
studied in \cite{Bento-etal-20}. Begley et al. present a new approach to
explaining fairness in machine learning, based on the Shapley value paradigm.
Antwarg et al. \cite{Antwarg-etal-20} study how to explain anomalies detected
by autoencoders using SHAP. The problem of explaining anomalies detected by
PCA is also considered by Takeishi \cite{Takeishi-19}. Bouneder et al.
\cite{Bouneder-etal-20} proposed X-SHAP which extends one of the
approximations of SHAP called the Kernel SHAP \cite{Lundberg-Lee-2017}. SHAP
is also applied to problems of explaining individual predictions when features
are dependent \cite{Aas-etal-2019} or when features are mixed
\cite{Redelmeier-etal-20}. SHAP has been used in real applications to explain
predictions of the black-box models, for example, it was used to rank failure
modes of reinforced concrete columns and to explains why a machine learning
model predicts a specific failure mode for a given sample
\cite{Mangalathu-etal-20}. It was also used in chemoinformatics and medicinal
chemistry \cite{Rodriguez-20}. An interesting application of SHAP in the
desirable interpretation of the machine learning-based model results for
identifying m7G sites in the gene expression analysis was proposed by Bi et
al. \cite{Bi-Xiang-etal-20}. Basic problems of SHAP are also analyzed by Kumar
et al. \cite{Kumar-etal-20}.

Many other interpretation methods, their analysis, and critical reviews can be
found also in survey papers
\cite{Adadi-Berrada-2018,Arrieta-etal-2019,Belle-Papantonis-2020,Carvalho-etal-2019,Das-Rad-20,Guidotti-2019,Liang-etal-2021,Rudin-2019,Xie-Ras-etal-2020}%
.

\section{Shapley values and the explanation model}

One of the most powerful approaches to explaining predictions of the black-box
machine learning models is the approach based on using the Shapley values
\cite{Shapley-1953} as a key concept in coalitional games. According to the
concept, the total gain of a game is distributed to players such that
desirable properties, including efficiency, symmetry, and linearity, are
fulfilled. In the framework of the machine learning, the gain can be viewed as
the machine learning model prediction or the model output, and a player is a
feature of input data. Hence, contributions of features to the model
prediction can be estimated by Shapley values. The $i$-th feature importance
is defined by the Shapley value%
\begin{equation}
\phi_{i}(f)=\phi_{i}=\sum_{S\subseteq N\backslash\{i\}}B(S,N)\left[  f\left(
S\cup\{i\}\right)  -f\left(  S\right)  \right]  , \label{SHAP_1}%
\end{equation}
where $f\left(  S\right)  $ is the characteristic function in terms of
coalitional games or the black-box model prediction under condition that a
subset $S$ of features are used as the corresponding input in terms of machine
learning; $N$ is the set of all features; $B(S,N)$ is defined as
\begin{equation}
B(S,N)=\frac{\left\vert S\right\vert !\left(  \left\vert N\right\vert
-\left\vert S\right\vert -1\right)  !}{\left\vert N\right\vert !}.
\end{equation}

It can be seen from the above expression that the Shapley value $\phi_{i}$ can
be regarded as the average contribution of the $i$-th feature across all
possible permutations of the feature set.

The Shapley value has the following important properties:

\textbf{Efficiency}. The total gain is distributed as $\sum_{k=0}^{m}\phi
_{k}=f(\mathbf{x})$.

\textbf{Symmetry}. If two players with numbers $i$ and $j$ make equal
contributions, i.e., $f\left(  S\cup\{i\}\right)  =f\left(  S\cup\{j\}\right)
$ for all subsets $S$ which contain neither $i$ nor $j$, then $\phi_{i}%
=\phi_{j}$.

\textbf{Dummy}. If a player makes zero contribution, i.e., $f\left(
S\cup\{j\}\right)  =f\left(  S\right)  $ for a player $j$ and all $S\subseteq
N\backslash\{j\}$, then $\phi_{j}=0$.

\textbf{Linearity}. A linear combination of multiple games $f_{1},...,f_{n}$,
represented as $f(S)=\sum_{k=1}^{n}c_{k}f_{k}(S)$, has gains derived from $f$:
$\phi_{i}(f)=\sum_{k=1}^{m}c_{k}\phi_{i}(f_{k})$ for every $i$.

Let us consider a machine learning problem. Suppose that there is a dataset
$\{(\mathbf{x}_{1},y_{1}),...,(\mathbf{x}_{n},y_{n})\}$ of $n$ points
$(\mathbf{x}_{i},y_{i})$, where $\mathbf{x}_{i}\in\mathcal{X}\subset
\mathbb{R}^{m}$ is a feature vector consisting of $m$ features, $y_{i}$ is the
observed output for the feature vector $\mathbf{x}_{i}$ such that $y_{i}%
\in\mathbb{R}$ in the regression problem and $y_{i}\in\{1,2,...,T\}$ in the
classification problem with $T$ classes. If a task is to interpret or to
explain the prediction from the model $f(\mathbf{x}^{\ast})$ at a local
feature vector $\mathbf{x}^{\ast}$, then the prediction $f(\mathbf{x}^{\ast})$
can be represented by using Shapley values as follows
\cite{Lundberg-Lee-2017,Strumbel-Kononenko-2010}:
\begin{equation}
f(\mathbf{x}^{\ast})=\phi_{0}+\sum_{j=0}^{m}\phi_{j}^{\ast},
\end{equation}
where $\phi_{0}=\mathbb{E}[f(\mathbf{x})]$, $\phi_{j}^{\ast}$ is the value
$\phi_{j}$ for the prediction $\mathbf{x}=\mathbf{x}^{\ast}$.

The above implies that the Shapley values explain the difference between the
prediction $f(\mathbf{x}^{\ast})$ and the global average prediction.

One of the crucial questions for implementing the SHAP method is how to remove
features from subset $N\backslash S$, i.e., how to fill input features from
subset $N\backslash S$ in order to get predictions $f\left(  S\right)  $ of
the black-box model. A detailed description of various ways for removing
features is presented by Covert at al. \cite{Covert-etal-20}. One of the ways
is simply by setting the removed features to zero
\cite{Petsiuk-etal-2018,Zeiler-Fergus-2014} or by setting them to user-defined
default values \cite{Ribeiro-etal-2016}. According to the way, features are
often replaced with their mean values. Another way removes feature by
replacing them with a sample from a conditional generative model
\cite{Yu-Lin-Yang-etal-18}. In the LIME method for tabular data, features are
replaced with independent draws from specific distributions
\cite{Covert-etal-20} such that each distribution depends on original feature
values. These are only a part of all ways for removing features.

\section{Modifications of SHAP}

\subsection{ER-SHAP}

In spite of many approaches to simplify SHAP, it is difficult to expect a
significant simplification from the above modifications of SHAP. Therefore, a
new approach is proposed for simplifying the SHAP method and for reducing
computational expenses for calculating the Shapley values. A key idea behind
the proposed approach is to apply a modification of the random subspace method
\cite{Ho-98} and to consider an ensemble of random SHAPs. The approach is very
similar to the random forests when an ensemble of randomly built decision
trees is used to get some average classification or regression measures.

\begin{figure}
[ptb]
\begin{center}
\includegraphics[
height=2.5201in,
width=3.9972in
]%
{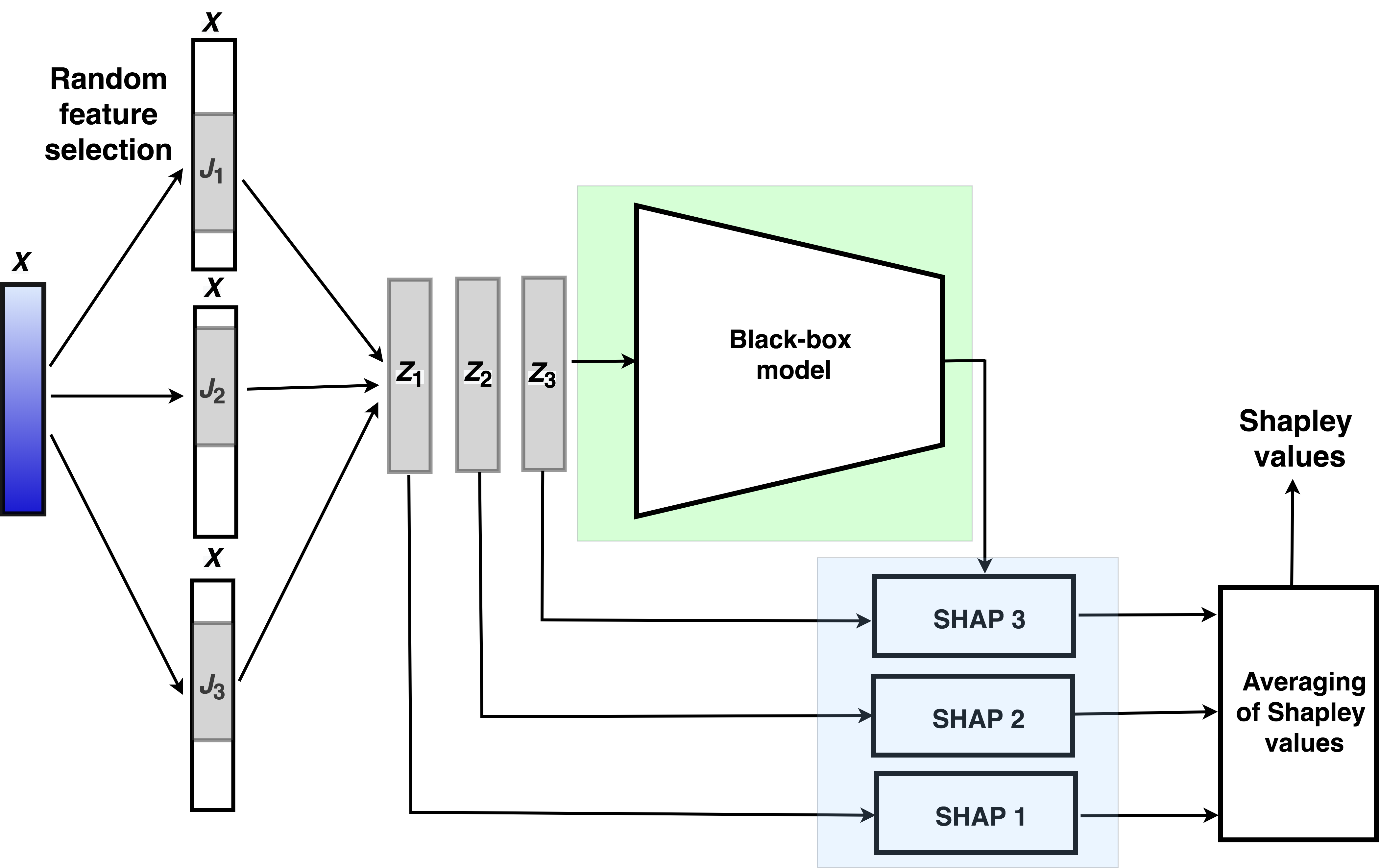}%
\caption{A scheme of the ER-SHAP}%
\label{fig:ER-SHAP-1}%
\end{center}
\end{figure}

Suppose that instance $\mathbf{x}\in\mathbb{R}^{m}$ has to be interpreted
under condition that the black-box model has been trained on the dataset
$D=\{(\mathbf{x}_{1},y_{1}),...,(\mathbf{x}_{n},y_{n})\}$. A general scheme of
the first approach called Ensemble of Random SHAPs (ER-SHAP) for case $N=3$ is
illustrated in Fig. \ref{fig:ER-SHAP-1}. ER-SHAP is iteratively constructed by
random selection of $t$ different features $N$ times. Value $t$ is a training
parameter. If we refer to random forests, then one of the heuristics is
$t\approx\sqrt{m}$. However, the optimal $t$ is obtained by considering many
its values. Suppose that indices of selected features at the $k$-th iteration
form the set $J_{k}=(i_{1},...,i_{t})$. The corresponding vector of $t$
features is regarded as an instance $\mathbf{z}_{k}=(x_{i_{1}},...,x_{i_{t}%
})\in\mathbb{R}^{t}$. Subsets of selected features with indices $J_{k}$ are
shown in Fig. \ref{fig:ER-SHAP-1} as successive features. However, this is
only a schematic illustration. Features are randomly selected in accordance
with the uniform distribution and can be located at arbitrary places of vector
$\mathbf{x}$.

As a result, we have a set of $N$ instances $\mathbf{z}_{1},...,\mathbf{z}%
_{N}$. The next step is to use the black-box model and SHAP to compute Shapley
values for every instance such that the subset $S_{k}=\{\phi_{i}^{(k)}:i\in
J_{k}\}$ of the Shapley values $\phi_{i}^{(k)}$ is produced for instance
$\mathbf{z}_{k}$. Repeating this procedure $N$ times, we get a set
$\mathcal{S}=\{S_{1},...,S_{N}\}$ of the Shapley values corresponding to all
$\mathbf{z}_{k}$, $k=1,...,N$, or all input subsets of features. Having set
$\mathcal{S}$, we can apply several combination rules to combining subsets
$S_{k}$ from $\mathcal{S}$. One of the simplest rules is based on averaging of
the Shapley values over all subsets $S_{k}$:
\begin{equation}
\phi_{i}=\frac{1}{N_{i}}\sum_{k:i\in J_{k}}\phi_{i}^{(k)},\ i=1,...,m,
\label{SHAP_10}%
\end{equation}
where $N_{i}$ is the number of the $i$-th feature selections among all
iterations, i.e. $N_{i}=\sum_{k:i\in J_{k}}1$.

It should be noted that the input of the black-box model has to have $m$
features. Therefore, for performing SHAPs with every $\mathbf{z}_{k}$, average
values of features over all dataset $D$ are used to fill $m-t$ remain features
though other methods \cite{Covert-etal-20} can be also used to fill these features.%

Algorithm 1 can be viewed as a formal scheme implementing ER-SHAP. It is
supposed that the black-box model has been already trained.

\begin{algorithm}
\caption{ER-SHAP}\label{alg:ER-SHAP-1}
\begin{algorithmic}
[1]\REQUIRE Training set $D$; point of interest $\mathbf{x}$; the number of
iterations $N$; the number of selected features $t$; the black-box model for
explaining $f(\mathbf{x})$
\ENSURE The Shapley values $S=\{\phi_{1}$,...,$\phi_{m}\}$
\FOR{$k=1$, $k\leq N$ }
\STATE Select randomly $t$ features from $\mathbf{x}$ and form the set $J_{k}$
of indices of randomly selected features $x_{i}$, $i\in J_{k}$
\STATE Use SHAP for computing $\phi_{i}^{(k)}$, $i\in J_{k}$ and form the set
$S_{k}=\{\phi_{i}^{(k)}:i\in J_{k}\}$
\ENDFOR
\STATE Combine sets $S_{k}$, $k=1,...,N$, to compute $S$, for example, by
using a simple averaging: $\phi_{i}=N_{i}^{-1}\sum_{k:i\in J_{k}}\phi
_{i}^{(k)}$, where $N_{i}=\sum_{k:i\in J_{k}}1$.
\end{algorithmic}
\end{algorithm}

If the number of subsets $S$ in the standard SHAP or the number of differences
$f\left(  S\cup\{i\}\right)  -f\left(  S\right)  $, which have to be computed
is $2^{m}$, then the number of the same differences in ER-SHAP is $N\cdot
2^{t}$. For comparison purposes, if we consider a dataset with $m=25$ and
$t=\sqrt{m}=5$, then $N$ can be taken $2^{25}/2^{5}=\allowbreak2^{20}$ in
order to make equal computational complexity of SHAP and ER-SHAP.

\subsection{ERW-SHAP}

\begin{figure}
[ptb]
\begin{center}
\includegraphics[
height=2.8575in,
width=4.4619in
]%
{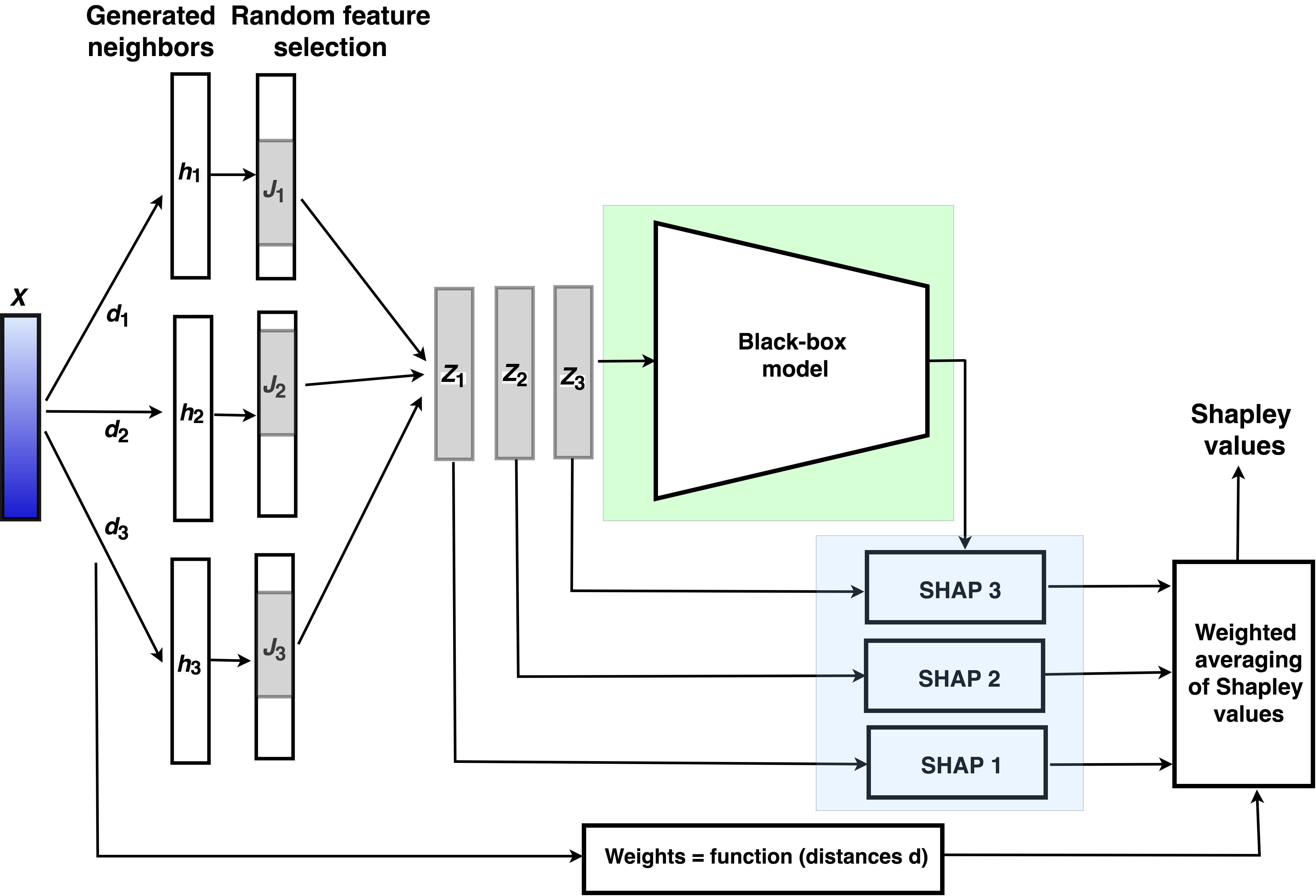}%
\caption{A scheme of the ERW-SHAP}%
\label{fig:ERW-SHAP}%
\end{center}
\end{figure}

The next algorithm is called the Ensemble of Random Weighted SHAPs (ERW-SHAP)
algorithm differs from ER-SHAP in the following parts. A general scheme is
shown in Fig. \ref{fig:ERW-SHAP}. First of all, $N$ points $\mathbf{h}%
_{1},...,\mathbf{h}_{N}$ are generated in the neighborhood of explained
instance $\mathbf{x}$. These points do not need to belong to the dataset $D$.
Then $t$ features are randomly selected from every $\mathbf{h}_{k}$, and they
produce instances $\mathbf{z}_{1},...,\mathbf{z}_{N}$. Moreover, the weight
$w_{k}$ of each instance $\mathbf{h}_{k}$ is defined as a function of the
distance $d_{k}$ between the explained instance $\mathbf{x}$ and the generated
neighbor $\mathbf{h}_{k}$. The weights are used to implement the weighted
average of the Shapley values. The final Shapley values are calculated now as
follows:
\begin{equation}
\phi_{i}=\frac{1}{W_{i}}\sum_{k:i\in J_{k}}w_{k}\phi_{i}^{(k)},\ i=1,...,m,
\label{SHAP_20}%
\end{equation}
where $W_{i}=\sum_{k:i\in J_{k}}w_{k}$.

On the one hand, using these changes of ER-SHAP, we implement an idea of some
kind of diversity of SHAPs to make the randomly selected feature vectors more
independent. On the other hand, the approach is similar to the LIME method
where the analyzed instance is perturbed in order to build an approximating
linear model around the instance to be explained. The diversity of SHAPs is a
very important peculiarity of the proposed ER-SHAP. It prevents SHAP from the
situation when a rule for filling the removed features produces features
coinciding with the explained instance features. In this case, the Shapley
values are incorrectly computed. The use of generated neighbors allows us to
avoid this case and to get more accurate results.%

Algorithm implementing ERW-SHAP differs from the similar Algorithm 1
implementing EW-SHAP only in two lines. First, after line 1 or before line 2,
the line indicating how to generate neighbors has to be inserted. Second, line
5 (combination of the Shapley values) is replaced with expression
(\ref{SHAP_20}).

\subsection{ER-SHAP-RF}

In order to control the process of the random feature selection, it is
reasonable to choose features for producing $\mathbf{z}_{1},...,\mathbf{z}%
_{N}$ in accordance with some probability distribution different from the
uniform distribution, which would take into account the preliminary importance
of features. The intuition behind this modification is to reduce the selection
of unimportant features which do not impact on the black-box prediction
corresponding to $\mathbf{x}$ a priori.\textbf{ }

One of the ways to implement this control is to compute the preliminary
feature importance by means of the random forest. Although it is known that
the random forest does not always give acceptable results related to the
feature selection problem, the proposed approach does not have this drawback
because we propose to train the random forest on instances generated in the
neighborhood of explained instance $\mathbf{x}$. The next algorithm is called
the Ensemble of Random SHAPs generated by the Random Forest (ER-SHAP-RF)
algorithm. The random forest plays a role of the important feature selection
model. It can be also viewed as some kind pre-training for important features.
The idea to train the random forest on generated neighbors allows us to
implement an preliminary explanation method. It should be noted that the
random forest is not a unique model for selecting important features. There
are many methods \cite{Speiser-etal-19}, which could be used for solving this
task. We use the random forest as one of the popular and simple methods having
a few parameters. In the same way, the linear regression model could be used
instead of the random forest. The random forest can be used as a explanation
model by applying an approach proposed by Sagi and Rokach
\cite{Sagi-Rokach-20} based on a scalable method for transforming a decision
forest into a single decision tree which is interpretable.

The LIME method can be also applied to get the probability distribution of
features. In the case of its use, normalized absolute values of linear
regression coefficients can be regarded as the probability distribution of features.

For solving the feature selection task by random forests, we use the
well-known simple method \cite{Breiman-2001}. According to this method, for
every tree from the random forest, we compute how much the impurity is
decreased by a feature. The more the feature decreases the impurity, the more
important the feature is. The impurity decreasing is averaged across all trees
in the random forest, and the obtained value corresponds to the final
importance of the feature.

The proposed approach may leads to small probabilities of unimportant
features. However, it does not mean that these features will not selected for
using in explanation by means of SHAP. They have a smaller chance to be
selected under condition that their probabilities are not equal to zero. This
implies that the classification or regression models for constructing the
probability distribution $P$ should not provide some sparse predictions like
the Lasso because only a small part of features in this case will take part in explanation.

\begin{figure}
[ptb]
\begin{center}
\includegraphics[
height=3.4489in,
width=3.8259in
]%
{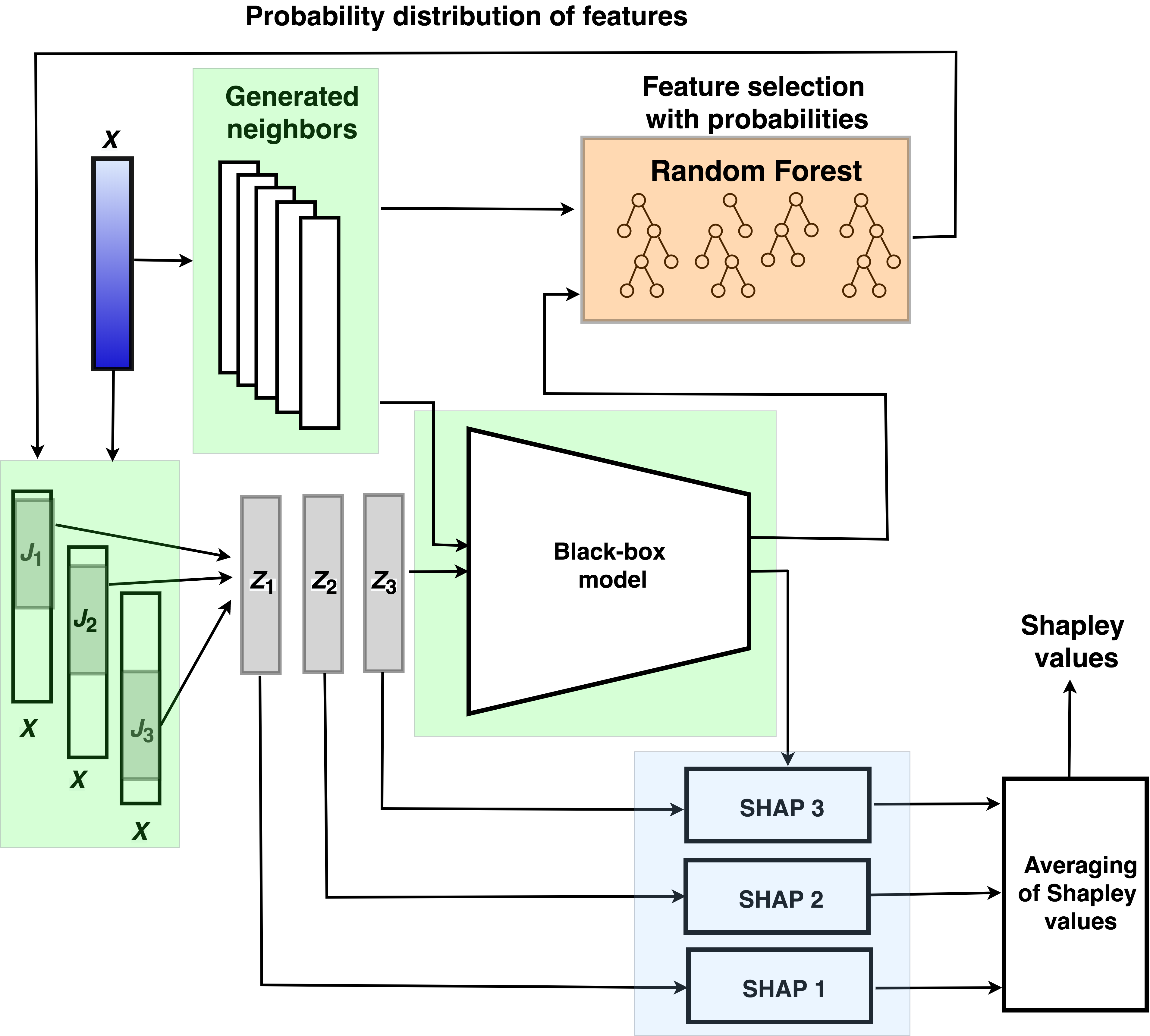}%
\caption{A scheme of the ER-SHAP-RF}%
\label{fig:ER-SHAP-RF}%
\end{center}
\end{figure}

A general scheme of ER-SHAP-RF is shown in Fig. \ref{fig:ER-SHAP-RF} where a
number, say $M$, of neighbors $\mathbf{h}_{1},...,\mathbf{h}_{M}$ are
generated around the instance $\mathbf{x}$ to be explained. Every generated
neighbor $\mathbf{h}_{j}$ is fed into the black-box model to get its class
label $y_{j}^{\ast}$. It should be noted that the training instances can be
taken as neighbors. However, they should be classified by using the black-box
model in order to take into account this model in explanation.

Having the obtained training with points $(\mathbf{h}_{j},y_{j}^{\ast})$, we
train the random forest which provides the feature importance measure in the
form of the probability distribution $P=(p_{1},...,p_{m})$. The distribution
$P$ is used to select features from instance $\mathbf{x}$ for constructing the
vectors $\mathbf{z}_{1},...,\mathbf{z}_{N}$, namely, $t$ features are selected
from $\mathbf{x}$ with replacement $N$ times in accordance with the
distribution $P$. SHAPs are used to find the Shapley values of vectors
$\mathbf{z}_{1},...,\mathbf{z}_{N}$. They are combined similarly to ERW-SHAP
by means of averaging as follows:
\begin{equation}
\phi_{i}=\frac{1}{N_{i}}\sum_{k:i\in J_{k}}\phi_{i}^{(k)},\ i=1,...,m.
\label{SHAP_28}%
\end{equation}

It is important that the number $N_{i}$ of the $i$-th feature selections among
all iterations $N$ is used instead of $N$.

The random forest should be built with a large depth of trees and with a small
number of trees in order to avoid a rather sparse probability distribution of
features when a large part of probabilities will be equal to zero or close to
zero. Another way for avoiding small probabilities of features is to apply
calibration methods and to recalculate the obtained probabilities, for
example, by using the temperature scaling as the simplest extension of Platt
scaling \cite{Chuan-etal-17}:%
\begin{equation}
p_{k}^{\ast}=\frac{\exp\left(  p_{k}/T\right)  }{%
{\textstyle\sum\nolimits_{i=1}^{m}}
\exp\left(  p_{i}/T\right)  },\ k=1,...,m, \label{SHAP_30}%
\end{equation}
where $T$ is the temperature which controls the smoothness of the probability
distribution, but it does not change the maximum of the softmax function.%

Algorithm 2 implementing ER-SHAP-RW can be viewed as an extension of ER-SHAP.

\begin{algorithm}
\caption{ER-SHAP-RF}\label{alg:ER-SHAP-RF}
\begin{algorithmic}
[1]\REQUIRE Training set $D$; point of interest $\mathbf{x}$; the number of
iterations $N$; the number of selected features $t$; the black-box model for
explaining $f(\mathbf{x})$; parameters of the random forest (the number and
depth of trees, number of instances for building trees)
\ENSURE The Shapley values $S=\{\phi_{1}$,...,$\phi_{m}\}$
\STATE Generate $M$ instances $\mathbf{h}_{1},...,\mathbf{h}_{M}$ which are
from the neighborhood of $\mathbf{x}$ or from the whole training set
\STATE Compute the class label $y_{j}^{\ast}=f(\mathbf{h}_{j})$ for every
generated instance by using the black-box model
\STATE Train the random forest on $(\mathbf{h}_{j},y_{j}^{\ast})$, $j=1,...,M$
\STATE Compute the probability distribution $P$ of features by using the
random forest
\FOR{$k=1$, $k\leq N$ }
\STATE Select randomly $t$ features from $\mathbf{x}$ in accordance with the
probability distribution $P$ and form the index set $J_{k}$ of features
\STATE Use SHAP for computing $\phi_{i}^{(k)}$, $i\in J_{k}$ and form the set
$S_{k}=\{\phi_{i}^{(k)}:i\in J_{k}\}$
\ENDFOR
\STATE Combine sets $S_{k}$, $k=1,...,N$, to compute $S$, for example, by
using a simple averaging: $\phi_{i}=N_{i}^{-1}\sum_{k:i\in J_{k}}\phi
_{i}^{(k)}$, where $N_{i}=\sum_{k:i\in J_{k}}1$.
\end{algorithmic}
\end{algorithm}

It is interesting to point out that the fourth algorithm can be also proposed,
which is represented as a combination of ERW-SHAP and ER-SHAP-RF. $N$ points
are generated for implementing diversity in accordance with ERW-SHAP, and $M$
points are generated for training the random forest in accordance with
ER-SHAP-RF and for computing the prior probability distribution $P$ of
features. Then the random features are selected not from the vector
$\mathbf{x}$ as it is done in ER-SHAP-RF, but from every vector $\mathbf{h}%
_{k}$ with the probability distribution $P$, $k=1,...,N$. However, this
algorithm is not studied because it can be regarded as the combination of
ERW-SHAP and ER-SHAP-RF, which are analyzed in detail.

\section{Numerical experiments}

\begin{figure}
[ptb]
\begin{center}
\includegraphics[
height=3.1989in,
width=5.585in
]%
{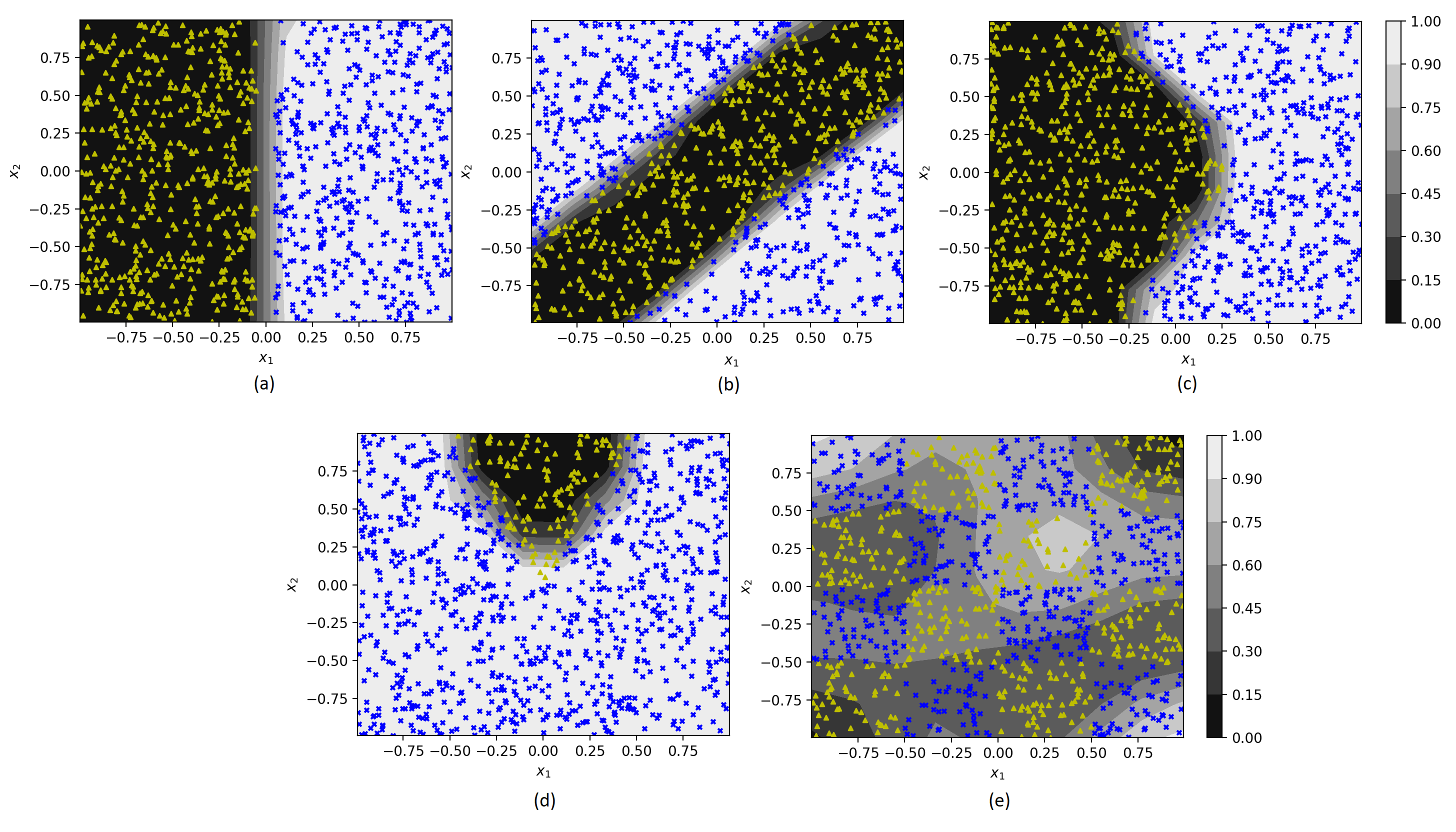}%
\caption{Five synthetic datasets and the boundaries between classes provided
by SVM}%
\label{f:initial_datasets_svm}%
\end{center}
\end{figure}

First, we consider several numerical examples for which training instances are
randomly generated. Each generated synthetic instance consists of 5 features.
Two features are generated as shown in Fig. \ref{f:initial_datasets_svm}, and
other features are uniformly generated in intervals $[-1,1]$. Each picture in
Fig. \ref{f:initial_datasets_svm} corresponds to a certain location of
instances of two classes such that the instances of classes $0$ and $1$ are
depicted by small triangles and crosses, respectively. This generation
corresponds to the case when the first two features may be important. These
features allows as to analyze the feature importance in accordance with the
data location and with the separating function. Other features are not
important, and they are used to generalize numerical experiments with
synthetic data.

Separating functions in Fig. \ref{f:initial_datasets_svm} are obtained by
means of SVM which can be regarded as the black-box model. It used the RBF
kernel whose parameter depends on a dataset trained. The SVM allows us to get
different separating functions by changing the kernel parameter. Fig.
\ref{f:initial_datasets_svm}(a) illustrates the linearly separating case. The
specific class area in the form of a stripe is shown in Fig.
\ref{f:initial_datasets_svm}(b). A saw-based separating function is used in
Fig. \ref{f:initial_datasets_svm}(c). The class area in the form of a wedge is
given in Fig. \ref{f:initial_datasets_svm}(d). A checkerboard with an attempt
of SVM to separate the checkerboard cages can be found in Fig.
\ref{f:initial_datasets_svm}(e). For every generated dataset from Fig.
\ref{f:initial_datasets_svm}, we compare SHAP with the proposed modifications.%

\textbf{Measures for comparison}: In order to compare the proposed
modifications with the original SHAP method, we use the concordance index $C$
of pairs, which is defined as the proportion of concordant pairs of the
Shapley values divided by the total number of possible evaluation pairs. Let
$\phi_{i}^{\ast}$ and $\phi_{i}$ be the Shapley values obtained by means of
the original SHAP method and one of its modifications (ER-SHAP, ERW-SHAP,
ER-SHAP-RF), respectively. Two pairs of the Shapley values $(\phi_{i},\phi
_{j})$ and $(\phi_{i}^{\ast},\phi_{j}^{\ast})$ are concordant if there hold
$(\phi_{i}>\phi_{j},\phi_{i}^{\ast}>\phi_{j}^{\ast})$ or $(\phi_{i}<\phi
_{j},\phi_{i}^{\ast}<\phi_{j}^{\ast})$. In contrast to the well-known C-index
in survival analysis, the introduced concordance index compares predictions
provided by two methods. If the index is close to 1, then the models provide
the same results. A motivation for the concordance index introduction is that
the Shapley values computed by using original SHAP and the proposed
modifications may be different. However, we are interesting in their
relationship. If the original SHAP method give the inequality $\phi_{i}^{\ast
}>\phi_{j}^{\ast}$ for some $i$ and $j$, then we are expecting to have
$\phi_{i}>\phi_{j}$ for the proposed method, but not equalities $\phi
_{i}^{\ast}=\phi_{i}$ and $\phi_{j}^{\ast}=\phi_{j}$. It should be noted that
original SHAP may provide incorrect results. Therefore, the introduced
concordance index should be viewed as a desirable measure under condition of
correct SHAP results. 

We use the Kernel SHAP \cite{Lundberg-Lee-2017} in numerical experiments and
compare obtained results with it. 

In spite of importance of the concordance index, we also use the normalized
Euclidean distance $E$ between vectors $(\phi_{1}^{\ast},...,\phi_{m}^{\ast})$
and $(\phi_{1},...,\phi_{m})$. The distance shows how the absolute Shapley
values of two methods are close to each other. It is important to take into
account that the Shapley values in the original SHAP method satisfy the
efficiency property when $\phi_{1}^{\ast}+...+\phi_{m}^{\ast}=f(\mathbf{x}%
)-f(\varnothing)$. This property is not fulfilled for modifications because
they do not enumerate all subsets of features. Therefore, in order to consider
the Shapley values in the same scale, all values $\phi_{i}$ and $\phi
_{i}^{\ast}$ are normalized to be in interval $[0,1]$.

\subsection{ER-SHAP}

\begin{figure}
[ptb]
\begin{center}
\includegraphics[
height=1.6965in,
width=6.285in
]%
{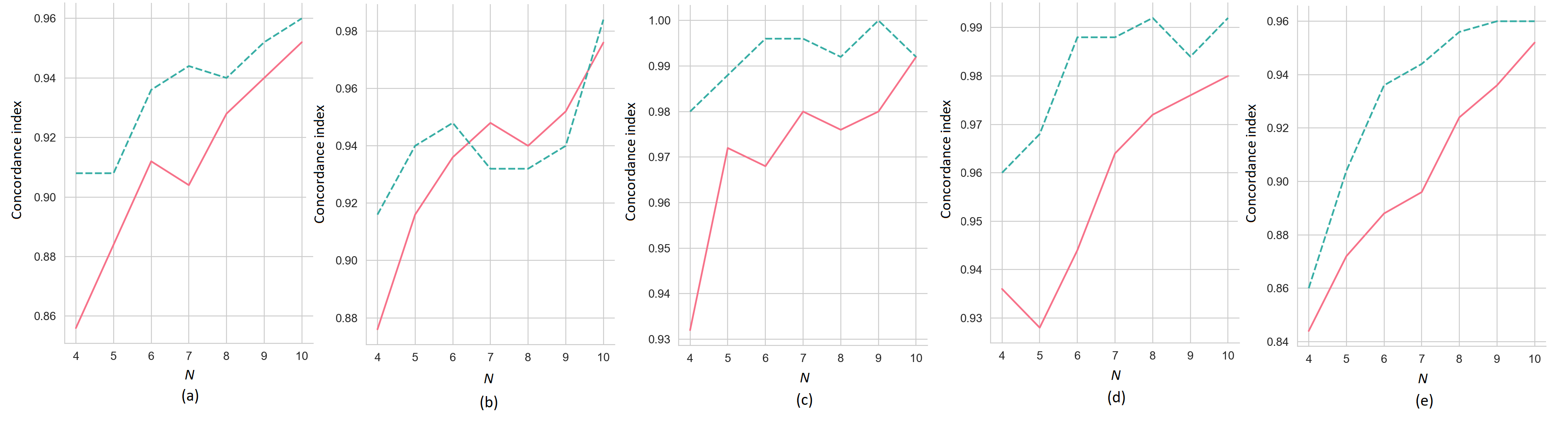}%
\caption{Concordance indices of ER-SHAP as functions of the number of
iterations $N$ for $t=2$ (the solid line) and $3$ (the dashed line) and for
five datasets and trained SVMs}%
\label{f:ER-SHAP-S-Concordance}%
\end{center}
\end{figure}

First, we consider results of numerical experiments obtained by means of
ER-SHAP with the SVM as a black-box model trained on the datasets shown in
Fig. \ref{f:initial_datasets_svm}. The explained instance for experiments has
all identical features which are equal to $0.25$. The concordance indices of
ER-SHAP as functions of the number of iterations $N$ for the numbers of
selected features $t=2$ (the solid line) and $t=3$ (the dashed line) are
illustrated in Fig. \ref{f:ER-SHAP-S-Concordance}, where pictures (a)-(e)
correspond to pictures (a)-(e) shown in Fig. \ref{f:initial_datasets_svm}. It
can be seen from Fig. \ref{f:ER-SHAP-S-Concordance} that the concordance index
increases with $N$ on average. This implies that ER-SHAP provides results
comparable with SHAP. It can be also seen from pictures that the concordance
index is significantly larger for $t=3$ in comparison with the case of $t=2$.
This observation is obvious because the large number of selected features in
each iteration brings the modification closer the original SHAP method. Though
one can see from Fig. \ref{f:ER-SHAP-S-Concordance} (b) that the case $t=2$
provides better concordance index by $N\geq7$.%

\begin{figure}
[ptb]
\begin{center}
\includegraphics[
height=1.8277in,
width=6.2996in
]%
{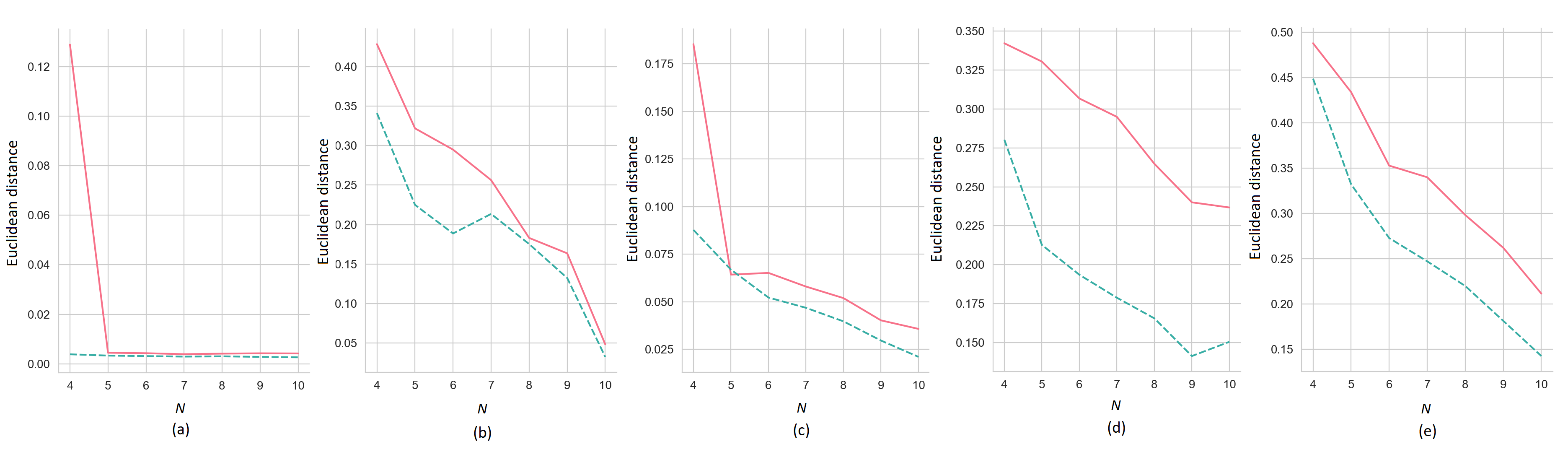}%
\caption{Euclidean distances between ER-SHAP and SHAP as functions of the
number of iterations $N$ for $t=2$ (the solid line) and $3$ (the dashed line)
and for five datasets and trained SVMs}%
\label{f:ER-SHAP-S-Euclid}%
\end{center}
\end{figure}

Fig. \ref{f:ER-SHAP-S-Euclid} illustrates how the Euclidean distances between
ER-SHAP and SHAP as functions of the number of iterations $N$ for $t=2$ (the
solid line) and $3$ (the dashed line) decrease with $N$. We again consider
five training sets shown in Fig. \ref{f:initial_datasets_svm}.%

\begin{figure}
[ptb]
\begin{center}
\includegraphics[
height=1.6733in,
width=6.3476in
]%
{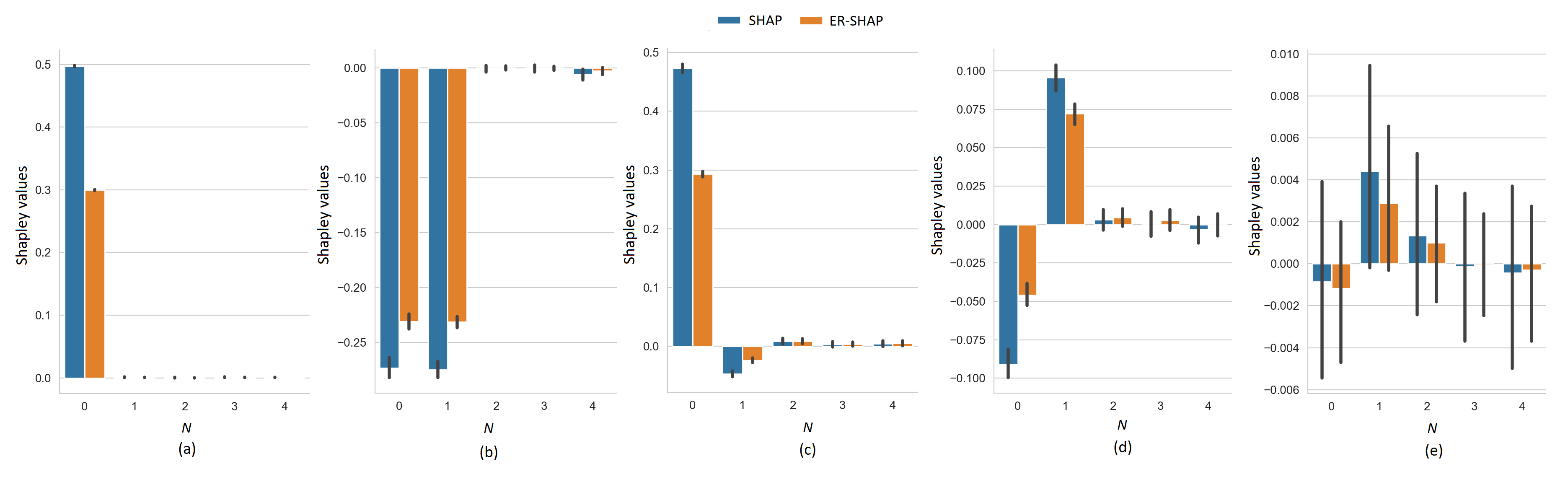}%
\caption{Shapley values obtained by means of SHAP and ER-SHAP as functions of
the number of iterations $N$ for five datasets and trained SVMs}%
\label{f:ER-SHAP-S-Contrib}%
\end{center}
\end{figure}

In order to explicitly illustrate how the Shapley values $\phi_{i}^{\ast}$ and
$\phi_{i}$ obtained by SHAP and ER-SHAP, respectively, are close to each
other, we show the Shapley values for all five cases in Fig.
\ref{f:ER-SHAP-S-Contrib}. It can be seen that despite the difference in
absolute values, the Shapley values indicate on the same important features.%

\subsection{ERW-SHAP}

To study ERW-SHAP, features of the explained instance are noised by using the
normal distribution of noise with the zero expectation and standard deviations
$0.01$ and $0.1$. Weights of generated instances $\mathbf{h}_{i}$ are defined
by
\begin{equation}
w_{i}=\exp\left(  -\left\Vert \mathbf{h}_{i}-\mathbf{x}\right\Vert
^{2}\right)  .
\end{equation}

\begin{figure}
[ptb]
\begin{center}
\includegraphics[
height=1.6381in,
width=5.8776in
]%
{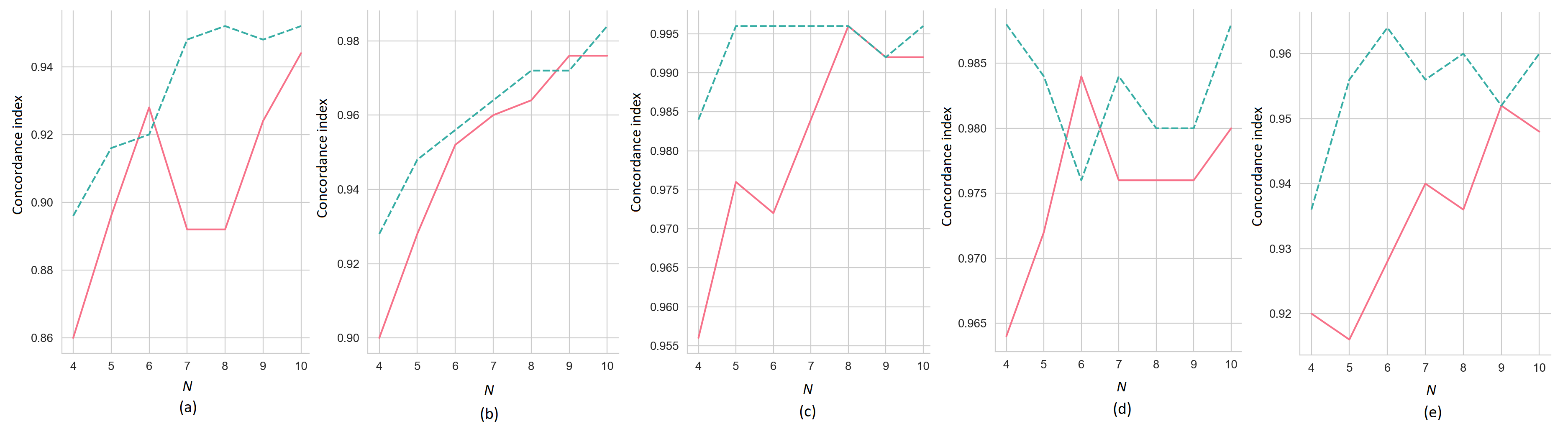}%
\caption{Concordance indices of ERW-SHAP as functions of $N$ for $t=2$ (the
solid line) and $3$ (the dashed line) and for five datasets and trained SVMs}%
\label{f:ERW-SHAP-S-Concordance}%
\end{center}
\end{figure}

We consider similar results of numerical experiments obtained by means of
ERW-SHAP with the SVM as a black-box model trained on the datasets shown in
Fig. \ref{f:initial_datasets_svm} with the same explained instance. The
concordance indices of ERW-SHAP as functions of $N$ for $t=2$ (the solid line)
and $t=3$ (the dashed line) are illustrated in Fig.
\ref{f:ERW-SHAP-S-Concordance}, where pictures (a)-(e) correspond to pictures
(a)-(e) shown in Fig. \ref{f:initial_datasets_svm}. The standard deviation of
the normal distribution generating noise is $0.01$. If we compare the
concordance indices for ERW-SHAP (Fig. \ref{f:ERW-SHAP-S-Concordance}) and for
ER-SHAP (Fig. \ref{f:ER-SHAP-S-Concordance}), then it is obvious that ERW-SHAP
provides better results in comparison with ERW-SHAP for most datasets.%

\begin{figure}
[ptb]
\begin{center}
\includegraphics[
height=1.6321in,
width=5.9076in
]%
{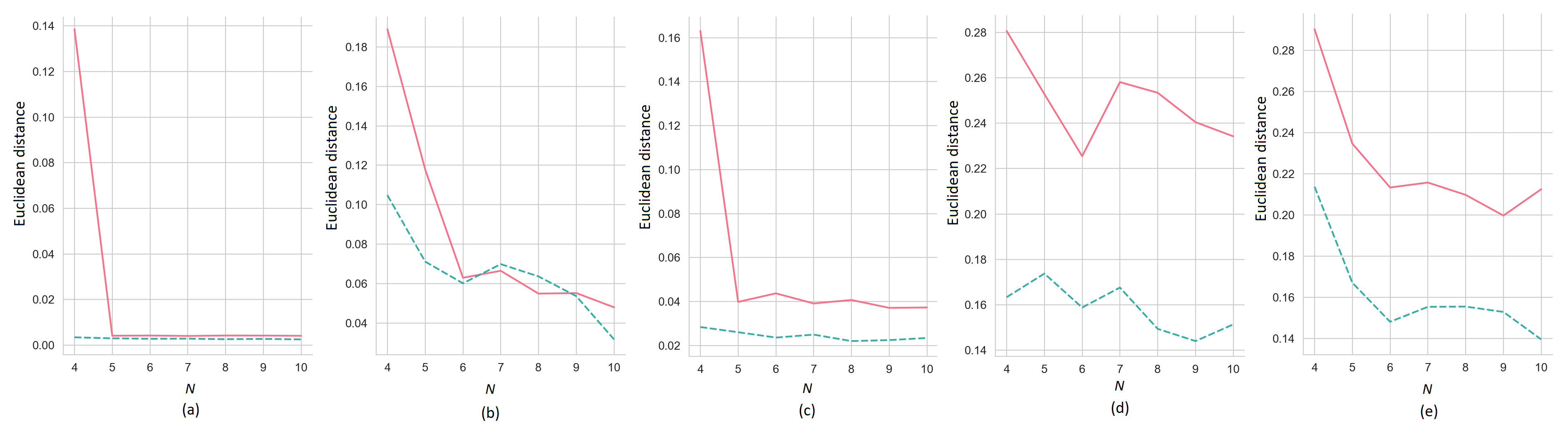}%
\caption{Euclidean distances between ERW-SHAP and SHAP as functions of $N$ for
$t=2$ (the solid line) and $3$ (the dashed line) and for five datasets and
trained SVMs}%
\label{f:ERW-SHAP-S-Euclid}%
\end{center}
\end{figure}

At the same time, the Euclidean distances between SHAP and ERW-SHAP slightly
differ from the same distances between SHAP and ER-SHAP. This follows from
Fig. \ref{f:ERW-SHAP-S-Euclid} where Euclidean distances between ERW-SHAP and
SHAP as functions of $N$ for $t=2$ (the solid line) and $3$ (the dashed line)
are presented for the above datasets.%

\begin{figure}
[ptb]
\begin{center}
\includegraphics[
height=1.7222in,
width=5.9068in
]%
{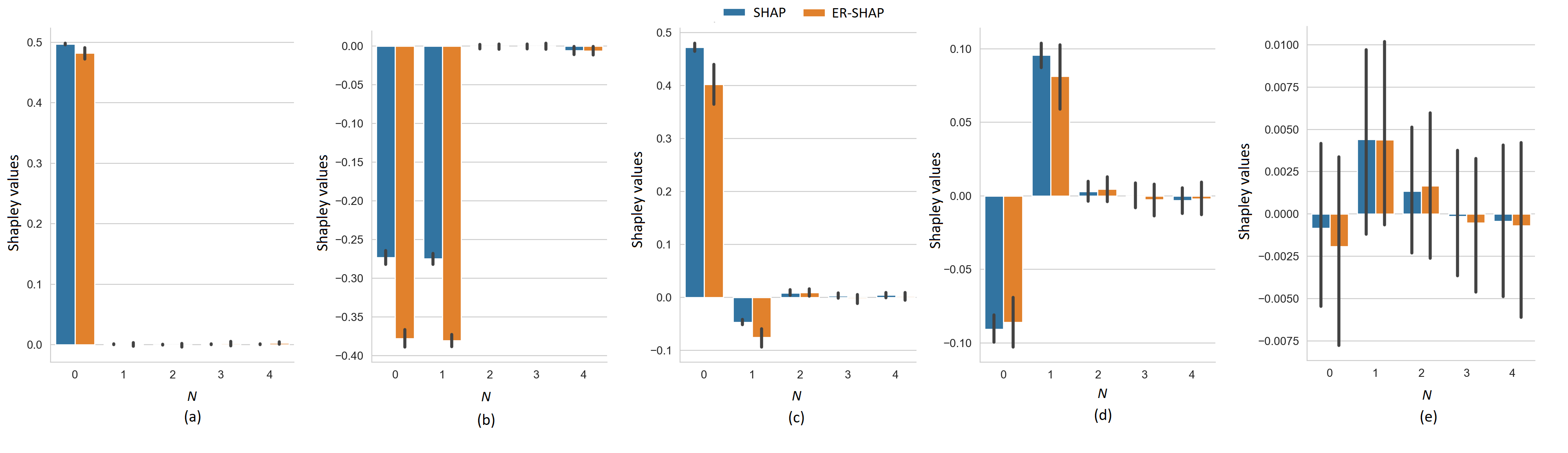}%
\caption{Shapley values obtained by means of SHAP and ERW-SHAP as functions of
the number of iterations $N$ for five datasets and trained SVMs under
condition of using the normal distribution of feature changes with the
standard deviation $0.1$}%
\label{f:ERW-SHAP-S-Contrib-1}%
\end{center}
\end{figure}

\begin{figure}
[ptb]
\begin{center}
\includegraphics[
height=1.7411in,
width=5.9342in
]%
{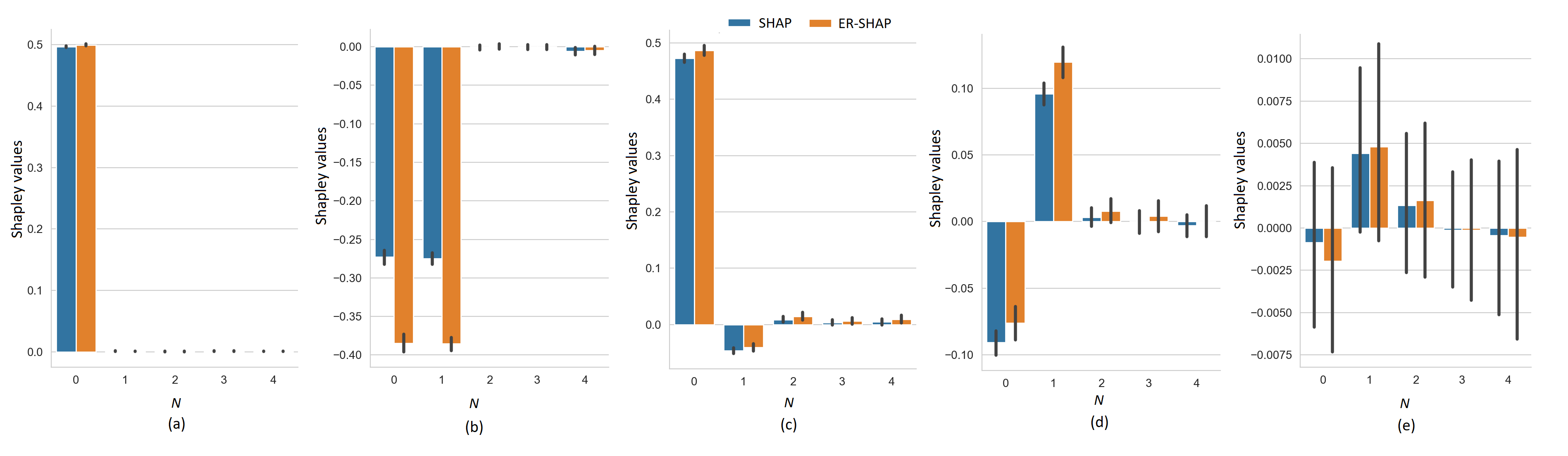}%
\caption{Shapley values obtained by means of SHAP and ERW-SHAP as functions of
the number of iterations $N$ for five datasets and trained SVMs under
condition of using the normal distribution of feature changes with the
standard deviation $0.01$}%
\label{f:ERW-SHAP-S-Contrib-01}%
\end{center}
\end{figure}

To illustrate how the Shapley values $\phi_{i}^{\ast}$ and $\phi_{i}$ obtained
by SHAP and ERW-SHAP, respectively, are close to each other, we show the
Shapley values for five cases in Fig. \ref{f:ERW-SHAP-S-Contrib-1} and in Fig.
\ref{f:ERW-SHAP-S-Contrib-01}. Figs. \ref{f:ERW-SHAP-S-Contrib-1} and
\ref{f:ERW-SHAP-S-Contrib-01} provide results under condition that the normal
distribution of the generated noise has the standard deviations $0.1$ and
$0.01$, respectively. We again observe that ERW-SHAP can be regarded as a good
approximation of SHAP because Shapley values of ERW-SHAP and SHAP are very
close to each other.%

%

\subsection{ER-SHAP-RF}

\begin{figure}
[ptb]
\begin{center}
\includegraphics[
height=1.6999in,
width=6.0783in
]%
{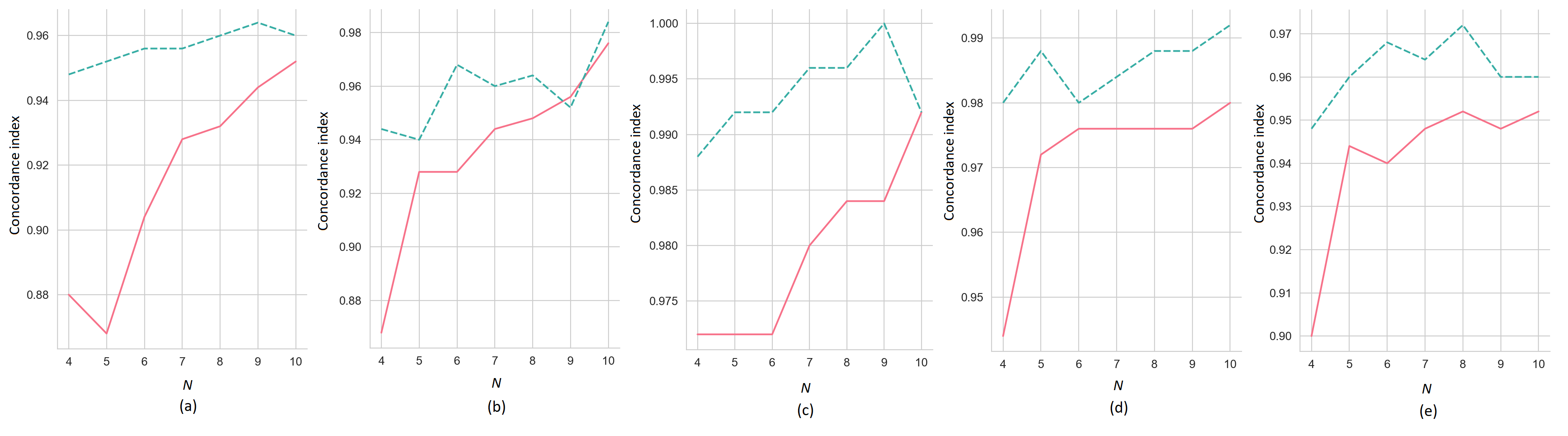}%
\caption{Concordance indices of ER-SHAP-RF as functions of $N$ for $t=2$ (the
solid line) and $3$ (the dashed line) and for five datasets and trained SVMs}%
\label{f:ER-SHAP-RF-S-Concordance}%
\end{center}
\end{figure}

\begin{figure}
[ptb]
\begin{center}
\includegraphics[
height=1.6845in,
width=6.1152in
]%
{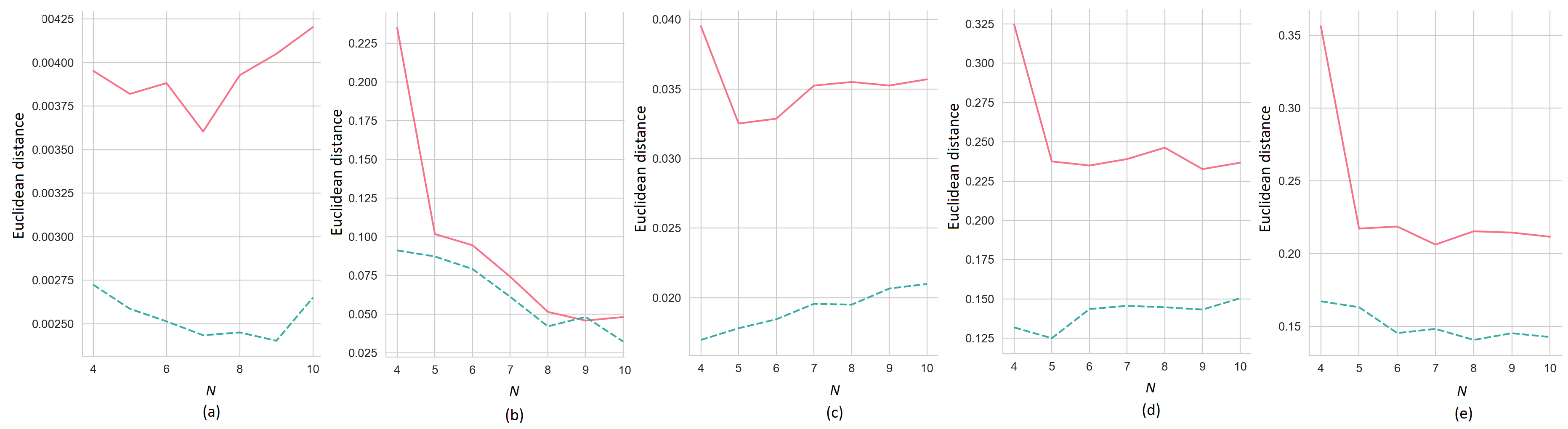}%
\caption{Euclidean distances between ER-SHAP-RF and SHAP as functions of $N$
for $t=2$ (the solid line) and $3$ (the dashed line) and for five datasets and
trained SVMs}%
\label{f:ER-SHAP-RF-S-Euclid}%
\end{center}
\end{figure}

\begin{figure}
[ptb]
\begin{center}
\includegraphics[
height=1.7599in,
width=6.116in
]%
{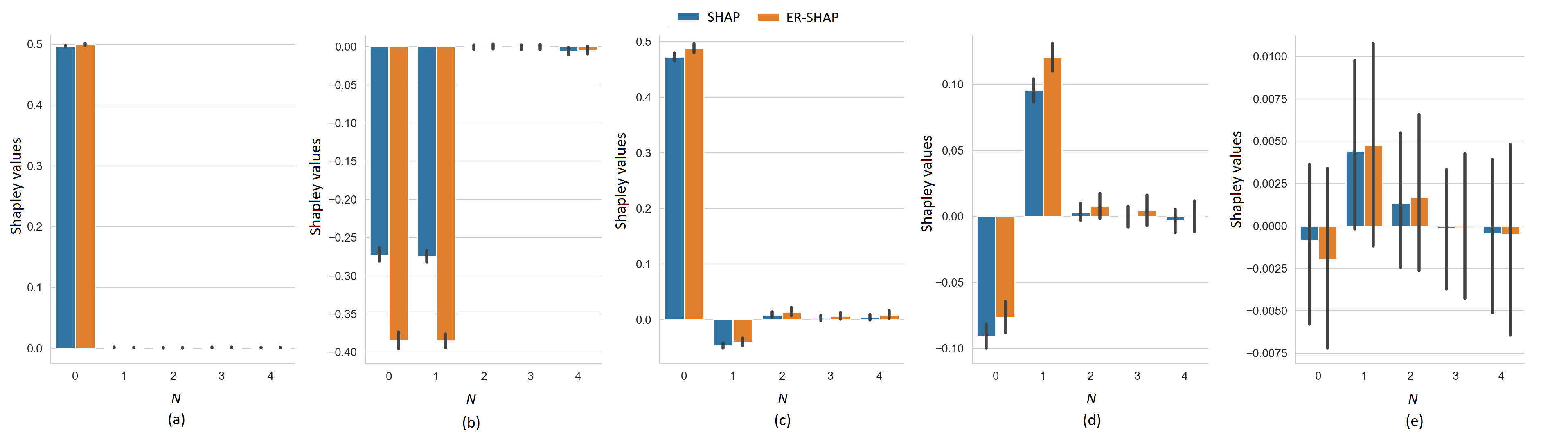}%
\caption{Shapley values obtained by means of SHAP and ER-SHAP-RF as functions
of $N$ for five datasets and trained SVMs}%
\label{f:ER-SHAP-RF-S-Contrib}%
\end{center}
\end{figure}

We again study the modification by using datasets shown in Fig.
\ref{f:initial_datasets_svm}. The result show that ER-SHAP-RF outperforms
ER-SHAP as well as ERW-SHAP for most dataset. Indeed, if we compare
concordance indices for ER-SHAP-RF (Fig. \ref{f:ER-SHAP-RF-S-Concordance})
with ERW-SHAP (Fig. \ref{f:ERW-SHAP-S-Concordance}) and ER-SHAP (Fig.
\ref{f:ER-SHAP-S-Concordance}), then we see that all examples provide better
results. In contrast to concordance indices, Euclidean distances shown in Fig.
\ref{f:ER-SHAP-RF-S-Euclid} demonstrate worse results. At the same time,
Shapley values given in Fig. \ref{f:ER-SHAP-RF-S-Contrib} almost coincide with
the corresponding values obtained by means of the ERW-SHAP (Fig.
\ref{f:ERW-SHAP-S-Contrib-01}). It should be noted that a more accurate tuning
of the random forest might provide outperforming results.%

%

%

\subsection{Boston Housing dataset}

Let us consider the real data called the Boston Housing dataset. It can be
obtained from the StatLib archive (http://lib.stat.cmu.edu/datasets/boston).
The Boston Housing dataset consists of 506 instances such that each instance is
described by 13 features.

\begin{figure}
[ptb]
\begin{center}
\includegraphics[
height=2.6472in,
width=3.4765in
]%
{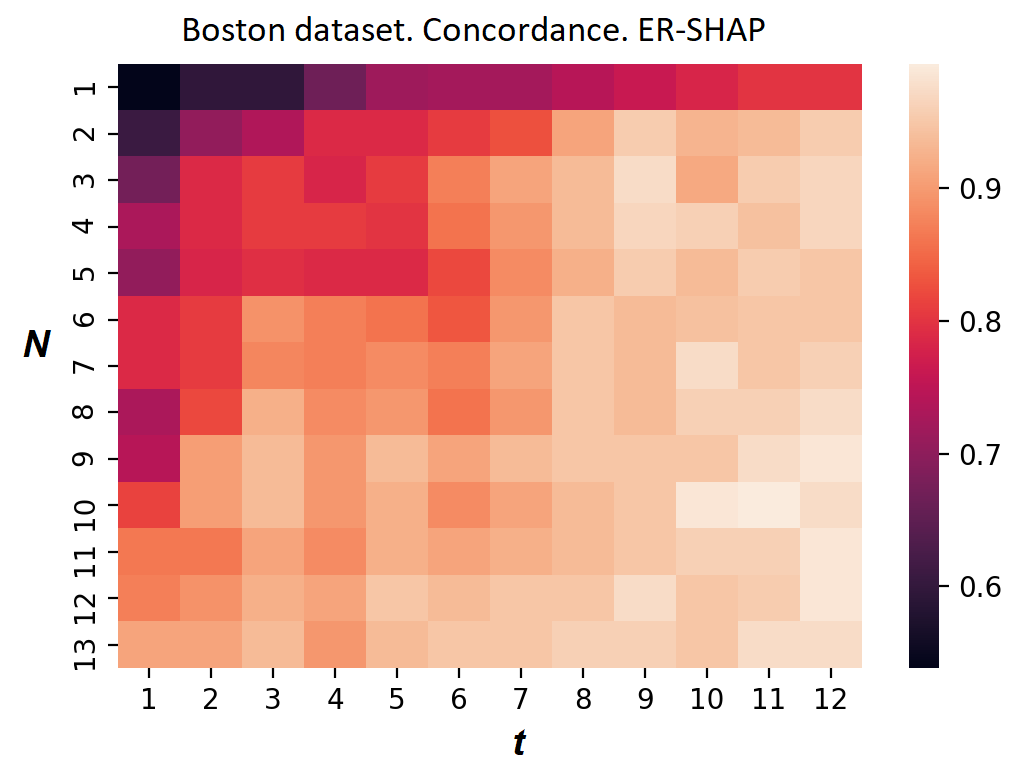}%
\caption{The heatmap reflecting the concordance index $C$ obtained by ER-SHAP
for the Boston Housing dataset}%
\label{f:heatmap_boston_1}%
\end{center}
\end{figure}

\begin{figure}
[ptb]
\begin{center}
\includegraphics[
height=2.6437in,
width=3.4662in
]%
{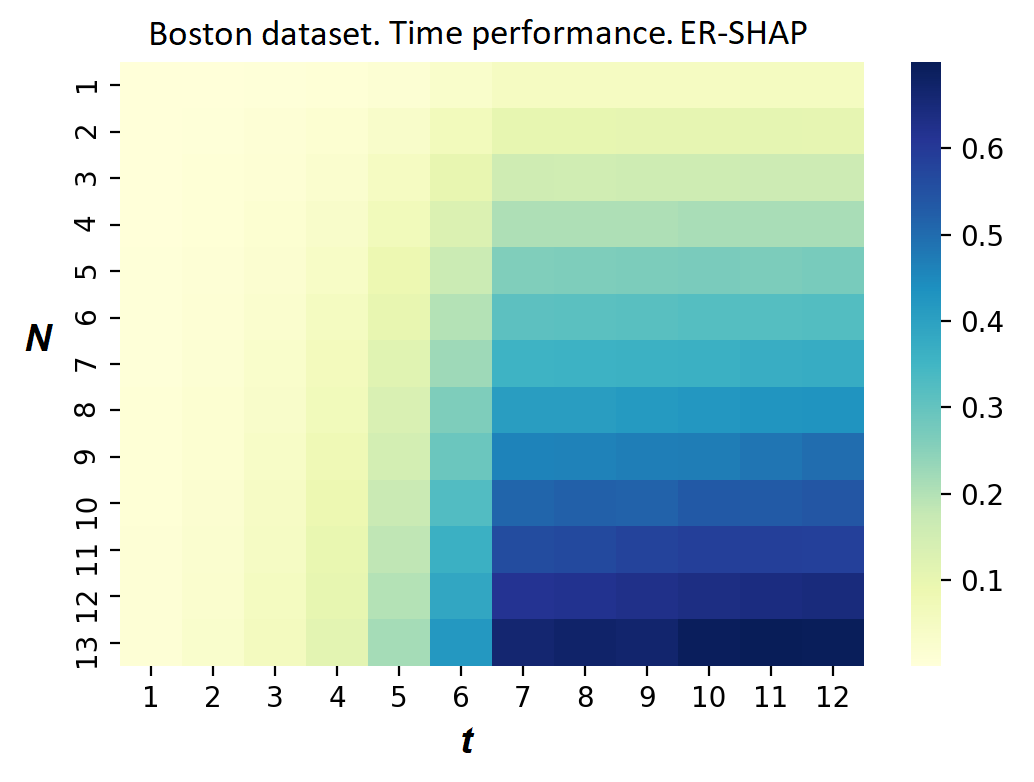}%
\caption{The heatmap illustrating the relationship between computation times
of SHAP and ER-SHAP for the Boston Housing dataset}%
\label{f:time_heatmap_boston_1}%
\end{center}
\end{figure}

The heatmap reflecting the concordance index of ER-SHAP for the Boston Housing
dataset is shown in Fig. \ref{f:heatmap_boston_1} Each element at position
$(i,j)$, where $i$ and $j$ are numbers of the row and column, respectively,
indicates the value the concordance index. Each row corresponds to the number
of iterations $N$, each column corresponds to the number of selected features
$t$. It can be seen from Fig. \ref{f:heatmap_boston_1} that the concordance
index increases with $N$ and $t$. This implies that ER-SHAP provides results
coinciding with SHAP by rather large numbers of iterations $N$. Fig.
\ref{f:time_heatmap_boston_1} illustrates how the computation time
$\tau_{\text{SHAP}}$ of SHAP exceeds the computation time $\tau
_{\text{ER-SHAP}}$ of ER-SHAP. The heatmap shows the ratio $\tau
_{\text{ER-SHAP}}/\tau_{\text{SHAP}}$. One can see from Fig.
\ref{f:time_heatmap_boston_1} a clear advantage of using ER-SHAP from the
computational point of view.

\begin{figure}
[ptb]
\begin{center}
\includegraphics[
height=2.6498in,
width=3.5379in
]%
{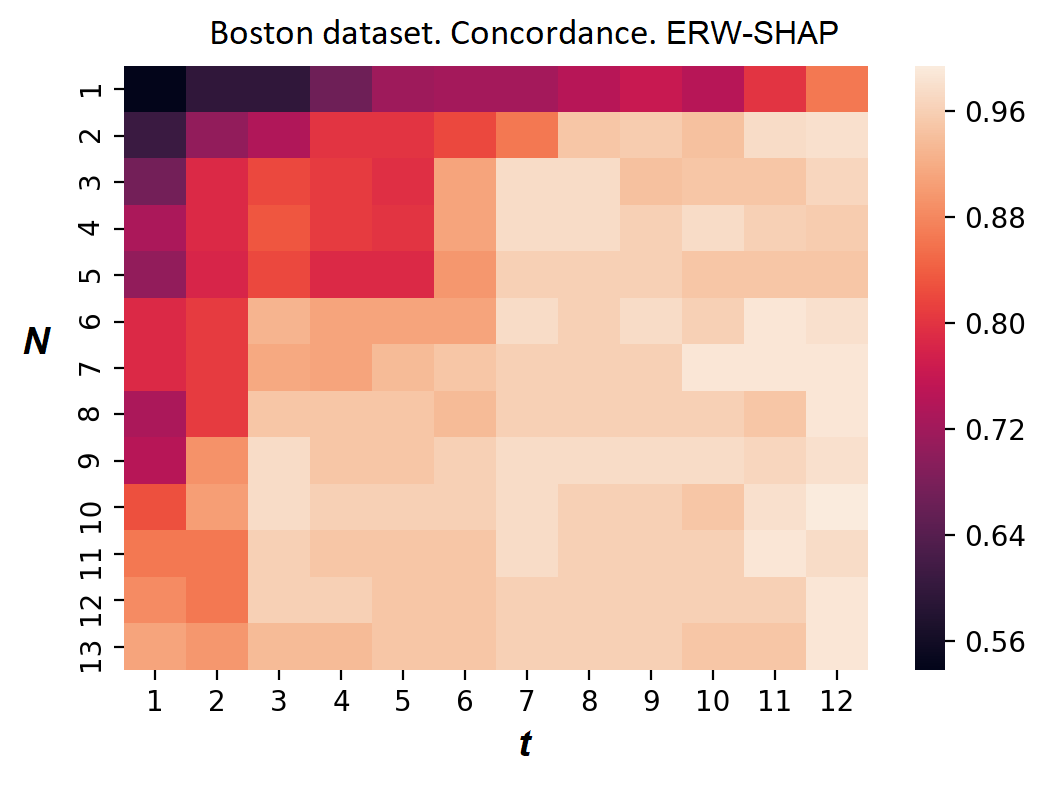}%
\caption{The heatmap reflecting the concordance index $C$ obtained by ERW-SHAP
for the Boston Housing dataset}%
\label{f:heatmap_boston_2}%
\end{center}
\end{figure}

Fig. \ref{f:heatmap_boston_2} shows the heatmap of the concordance index of
ERW-SHAP for the Boston Housing dataset. It is clearly seen from Fig.
\ref{f:heatmap_boston_2} that the introduction of weights and generated
instances significantly improves the approximation.%

%

%

\begin{figure}
[ptb]
\begin{center}
\includegraphics[
height=1.7351in,
width=5.1812in
]%
{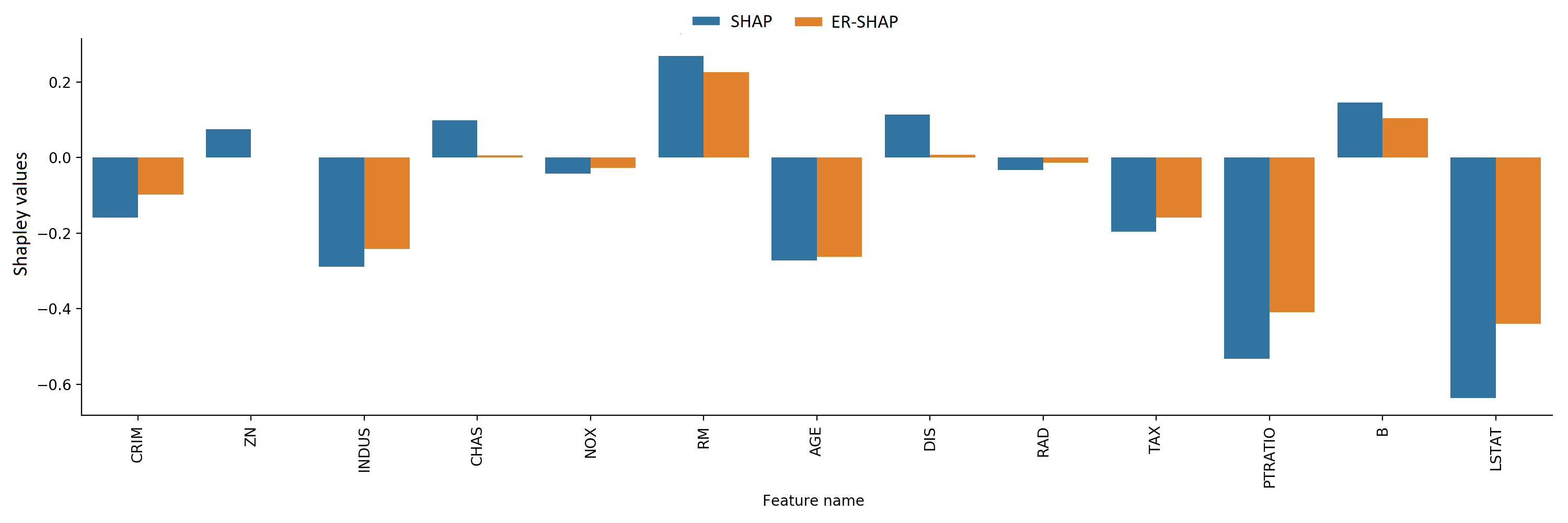}%
\caption{Shapley values obtained by means of SHAP and ER-SHAP as functions of
$N$ for the Boston Housing dataset}%
\label{f:boston_er_shapley}%
\end{center}
\end{figure}

\begin{figure}
[ptb]
\begin{center}
\includegraphics[
height=1.7908in,
width=5.3458in
]%
{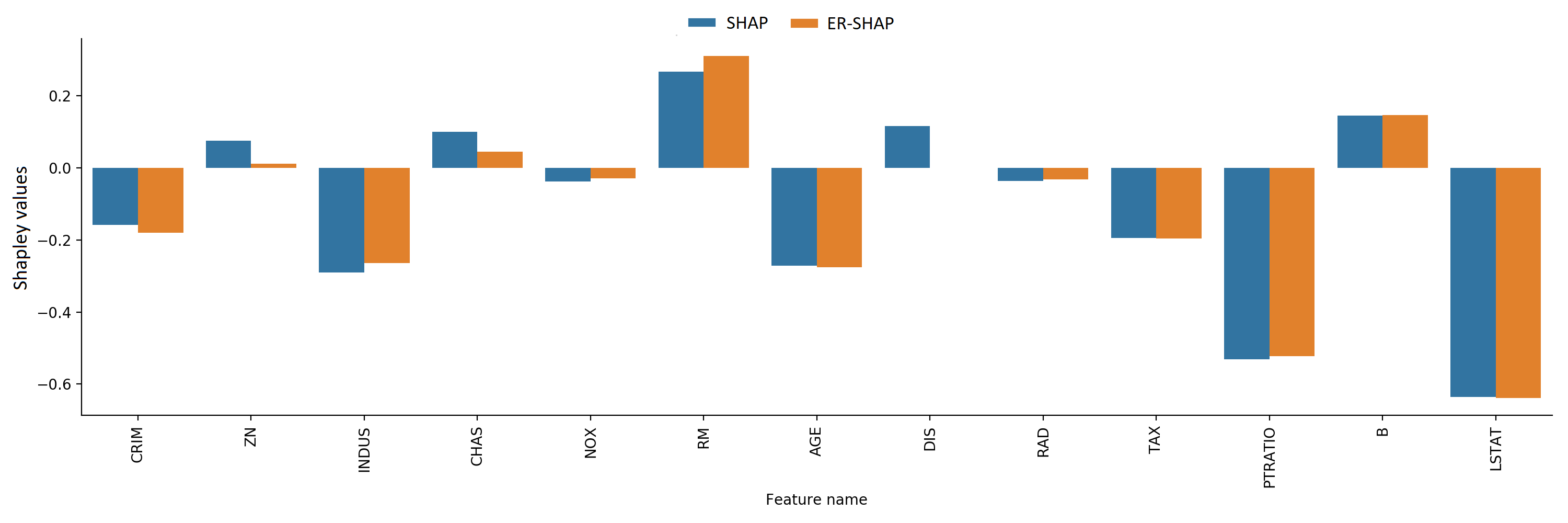}%
\caption{Shapley values obtained by means of SHAP and ERW-SHAP as functions of
$N$ for the Boston Housing dataset under condition of using the normal
distribution of feature changes with the standard deviation $0.1$}%
\label{f:boston_erw_shapley}%
\end{center}
\end{figure}

Shapley values obtained by means of ER-SHAP and SHAP as well as ERW-SHAP and
SHAP are shown in Figs. \ref{f:boston_er_shapley} and
\ref{f:boston_erw_shapley}, respectively. One can see from Figs.
\ref{f:boston_er_shapley} and \ref{f:boston_erw_shapley} that ERW-SHAP can be
viewed as a better approximation of SHAP because the corresponding bars almost
coincide as shown in Fig. \ref{f:boston_erw_shapley}. It should be noted that
Shapley values provided by ER-SHAP are also behaves like values of SHAP (see
Fig. \ref{f:boston_er_shapley}), but they do not coincide for the most
important features.%

%

\begin{figure}
[ptb]
\begin{center}
\includegraphics[
height=2.5728in,
width=3.525in
]%
{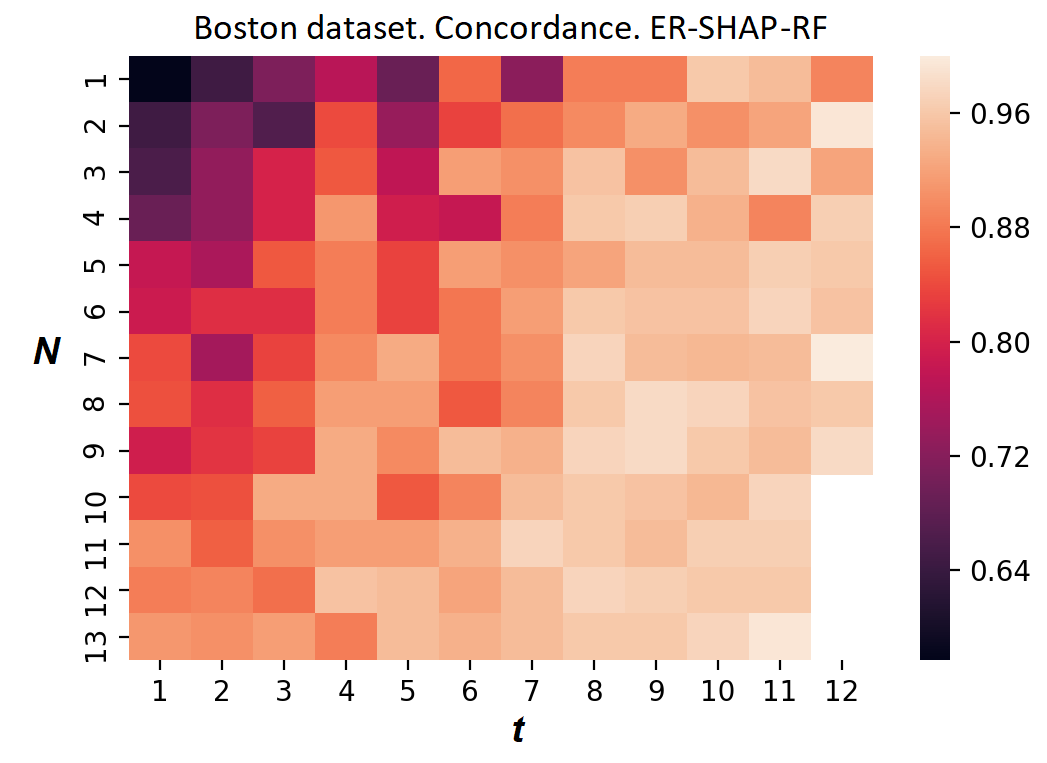}%
\caption{The heatmap reflecting the concordance index $C$ obtained by
ER-SHAP-RF for the Boston Housing dataset without using the temperature
scaling}%
\label{f:heatmap_boston_6}%
\end{center}
\end{figure}

\begin{figure}
[ptb]
\begin{center}
\includegraphics[
height=2.5755in,
width=3.525in
]%
{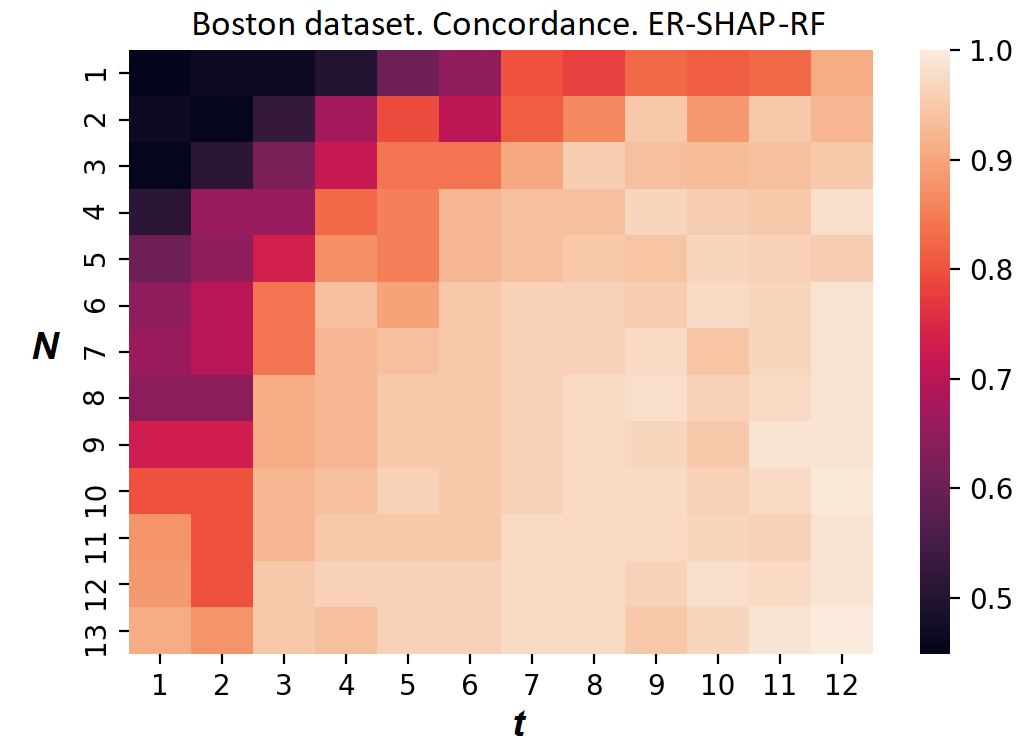}%
\caption{The heatmap reflecting the concordance index $C$ obtained by
ER-SHAP-RF for the Boston Housing dataset with using the temperature scaling}%
\label{f:heatmap_boston_7}%
\end{center}
\end{figure}

Figs. \ref{f:heatmap_boston_6} and \ref{f:heatmap_boston_7} illustrate
heatmaps of the concordance index of ER-SHAP-RF for the Boston Housing
dataset. They are obtained without using the temperature scaling in accordance
with (\ref{SHAP_30}) and with this calibration method, respectively. It is
interesting to observe from Figs. \ref{f:heatmap_boston_6} and
\ref{f:heatmap_boston_7} that the use of the calibration leads to a more
contrasting heatmap and to obvious improvement of the approximation quality.%

%

\subsection{Breast Cancer dataset}

The next real dataset is the Breast Cancer Wisconsin (Diagnostic). It can be
found in the well-known UCI Machine Learning Repository
(https://archive.ics.uci.edu). The Breast Cancer dataset contains 569 instances
such that each instance is described by 30 features. For classes of the breast
cancer diagnosis, the malignant and the benign are assigned by classes $0$ and
$1$, respectively. We consider the corresponding model in the framework of
regression with outcomes in the form of probabilities from 0 (malignant) to 1 (benign).

\begin{figure}
[ptb]
\begin{center}
\includegraphics[
height=2.77in,
width=3.6045in
]%
{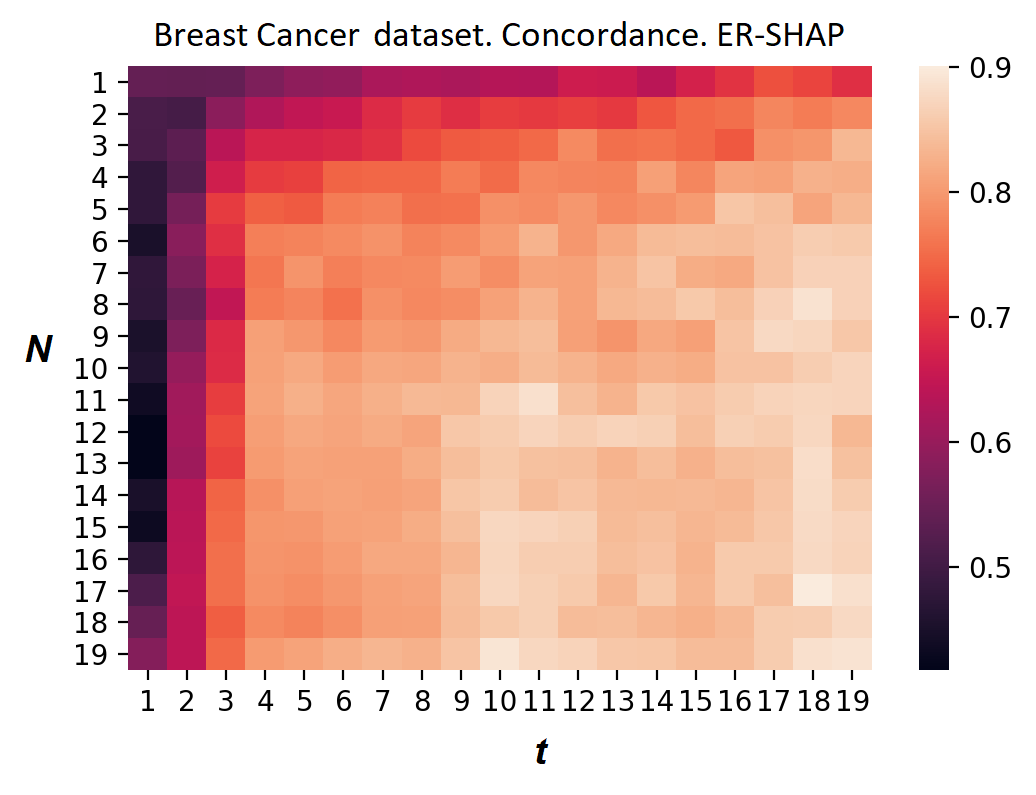}%
\caption{The heatmap reflecting the concordance index $C$ obtained by ER-SHAP
for the Breast Cancer dataset}%
\label{f:heatmap_breast_1}%
\end{center}
\end{figure}

\begin{figure}
[ptb]
\begin{center}
\includegraphics[
height=2.7501in,
width=3.6729in
]%
{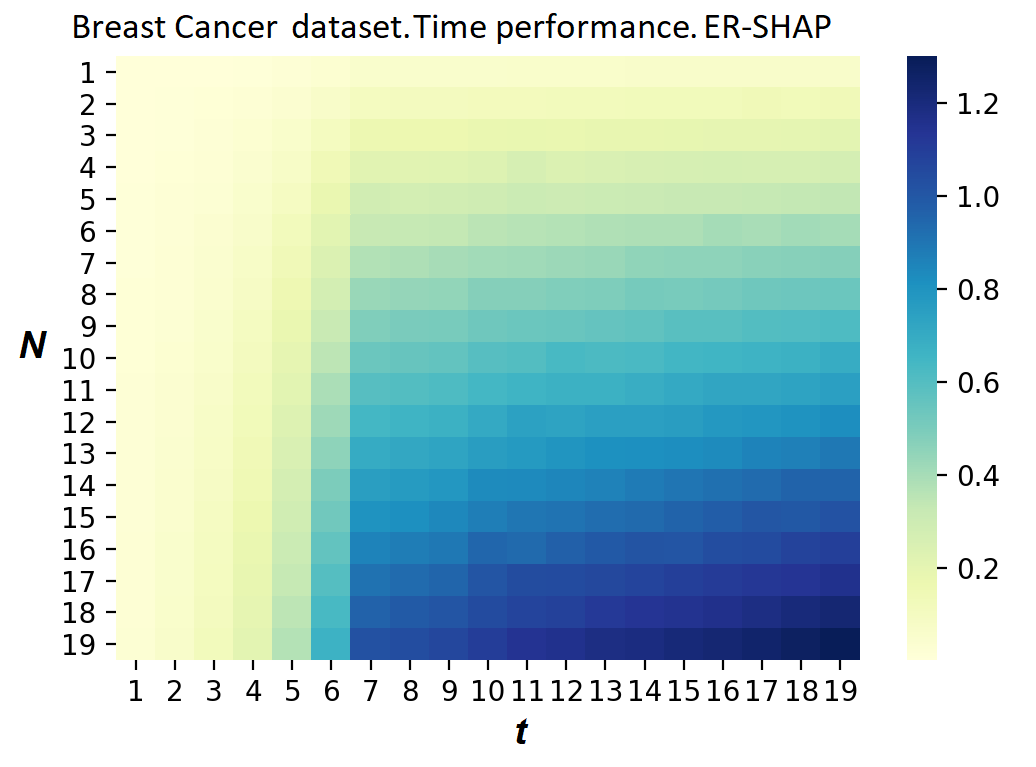}%
\caption{The heatmap illustrating the relationship between computation times
of SHAP and ER-SHAP for the Breast Cancer Housing dataset}%
\label{f:time_heatmap_breast_1}%
\end{center}
\end{figure}

Heatmaps given in Figs. \ref{f:heatmap_breast_1}-\ref{f:time_heatmap_breast_1}
are similar to the same heatmaps obtained for the Boston Housing dataset
(Figs. \ref{f:heatmap_boston_1}-\ref{f:time_heatmap_boston_1}). It is
interesting to observe from Fig. \ref{f:time_heatmap_breast_1} that there are
$N$ and $t$ such that the ratio $\tau_{\text{ER-SHAP}}/\tau_{\text{SHAP}}$ is
larger $1$. This implies that SHAP is computationally simpler in comparison
with ER-SHAP. However, these cases take place only for large values $N$ and
$t$.

\begin{figure}
[ptb]
\begin{center}
\includegraphics[
height=2.8669in,
width=3.736in
]%
{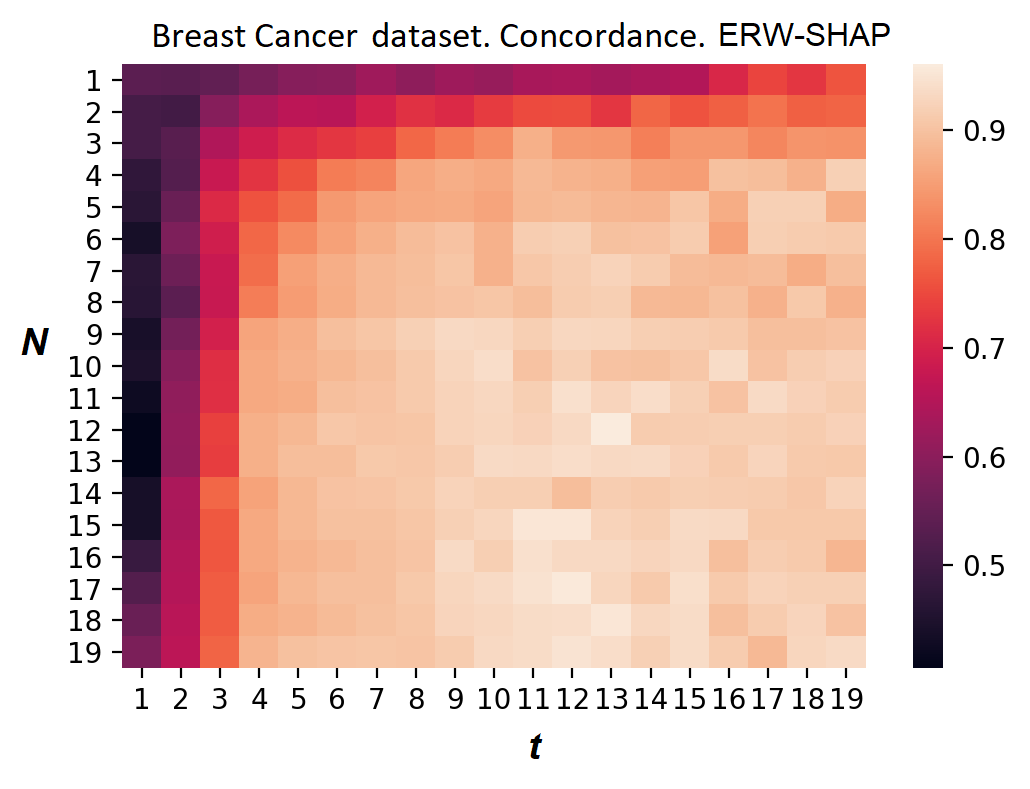}%
\caption{The heatmap reflecting the concordance index $C$ obtained by ERW-SHAP
for the Breast Cancer dataset}%
\label{f:heatmap_breast_2}%
\end{center}
\end{figure}

At first glance, it is difficult to evaluate from Fig.
\ref{f:heatmap_breast_2} whether ERW-SHAP provides better results than
ER-SHAP. Fig. \ref{f:heatmap_breast_2} shows the heatmap of the concordance
index of ERW-SHAP for the Breast Cancer dataset. However, we can be that the
legend in Fig. \ref{f:heatmap_breast_2} is changed in interval $[0.4,0.95]$,
whereas the legend in Fig. \ref{f:heatmap_breast_1} is changed in interval
$[0.4,0.9]$. This implies that ERW-SHAP outperforms ER-SHAP in this numerical example.%

%

%

\begin{figure}
[ptb]
\begin{center}
\includegraphics[
height=2.2042in,
width=6.5792in
]%
{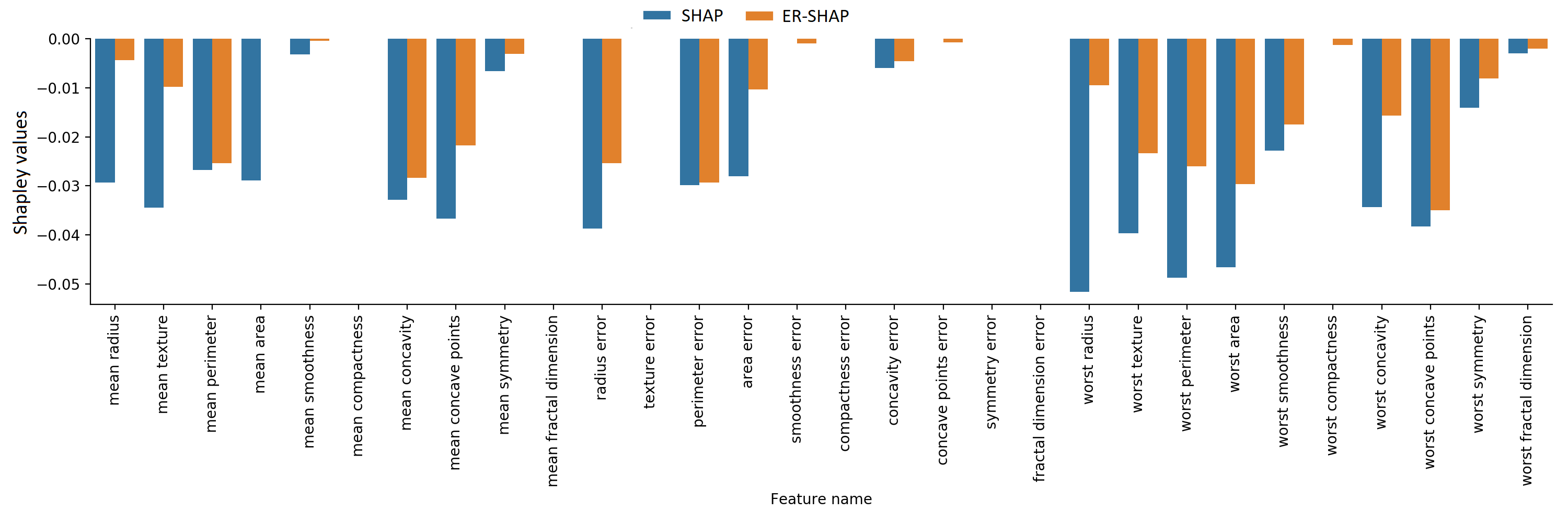}%
\caption{Shapley values obtained by means of SHAP and ER-SHAP as functions of
$N$ for the Breast Cancer dataset}%
\label{f:breast_er_shapley}%
\end{center}
\end{figure}

Shapley values for all features of the Breast Cancer dataset, which are
obtained by means of ER-SHAP and SHAP, are shown in Fig.
\ref{f:breast_er_shapley}. Similar values obtained by means of ERW-SHAP and
SHAP are shown in Fig. \ref{f:boston_erw_shapley}. One can see from Figs.
\ref{f:breast_er_shapley} and \ref{f:boston_erw_shapley} that the Shapley
values obtained by means of ERW-SHAP better approximate the SHAP Shapley
values. For example, if to look at the feature \textquotedblleft worst
radius\textquotedblright, which is important due to the original SHAP method,
then ER-SHAP provides the incorrect result whereas ERW-SHAP is totally
consistent with SHAP.%

%

\begin{figure}
[ptb]
\begin{center}
\includegraphics[
height=2.2102in,
width=6.5955in
]%
{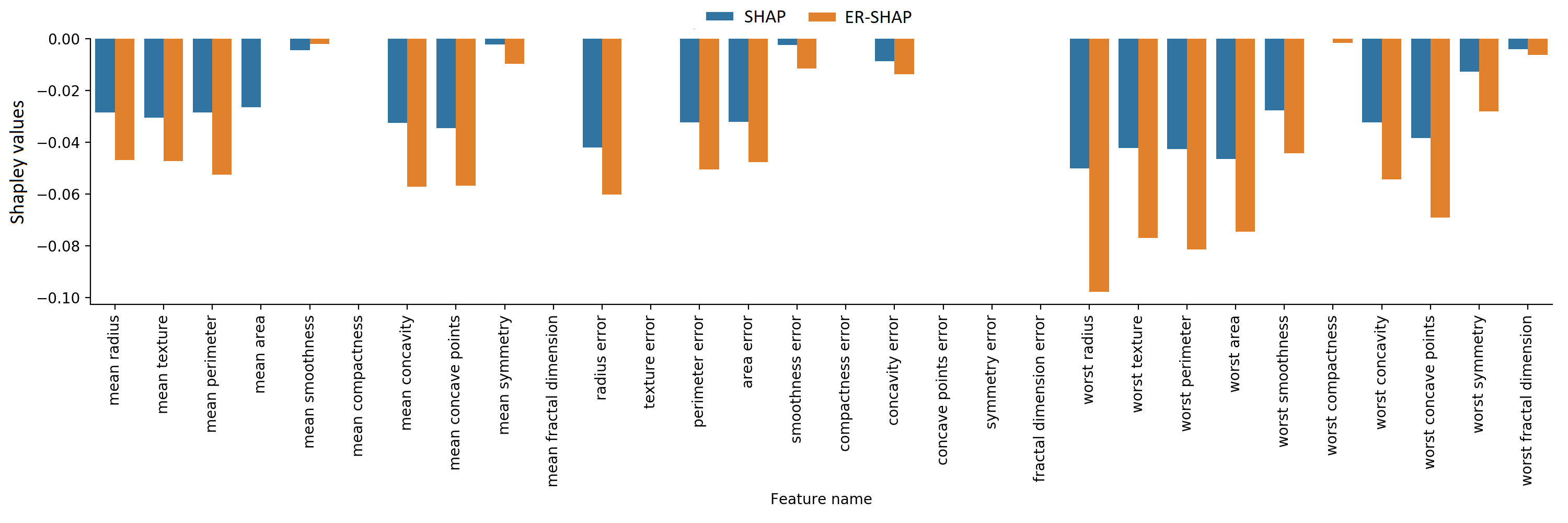}%
\caption{Shapley values obtained by means of SHAP and ERW-SHAP as functions of
$N$ for the Breast Cancer dataset under condition of using the normal
distribution of feature changes with the standard deviation $0.1$}%
\label{f:breast_erw_shapley}%
\end{center}
\end{figure}

\begin{figure}
[ptb]
\begin{center}
\includegraphics[
height=2.5935in,
width=3.4772in
]%
{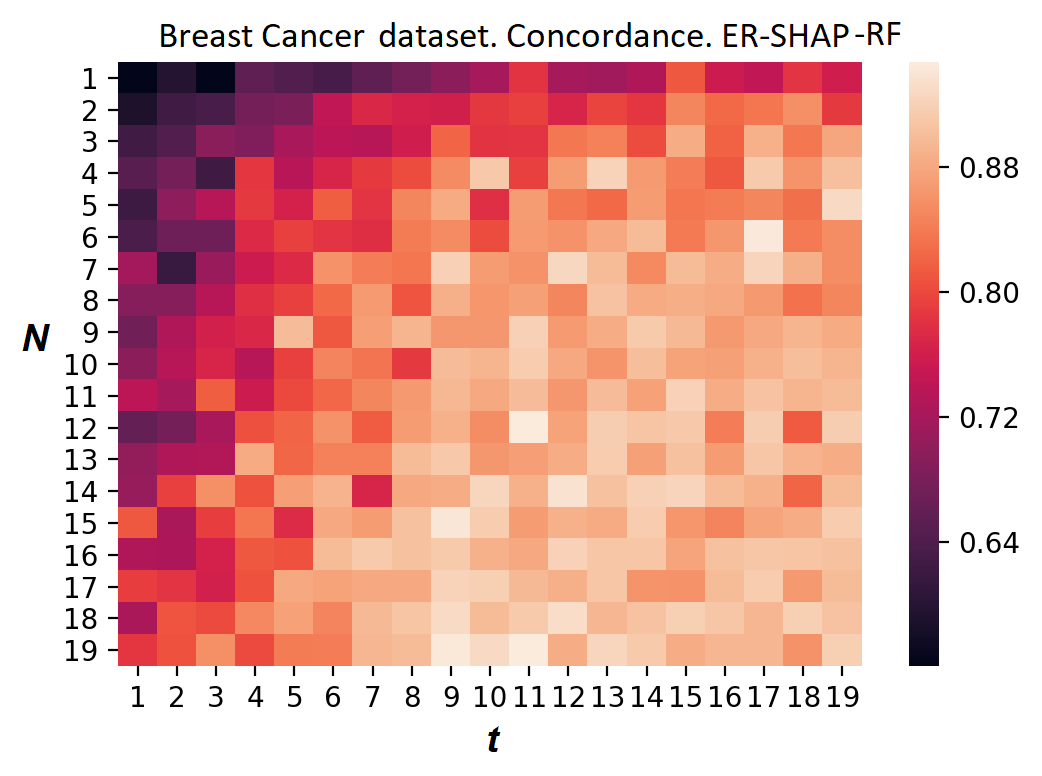}%
\caption{The heatmap reflecting the concordance index $C$ obtained by
ER-SHAP-RF for the Breast Cancer dataset without using the temperature
scaling}%
\label{f:heatmap_breast_6}%
\end{center}
\end{figure}

\begin{figure}
[ptb]
\begin{center}
\includegraphics[
height=2.6304in,
width=3.5241in
]%
{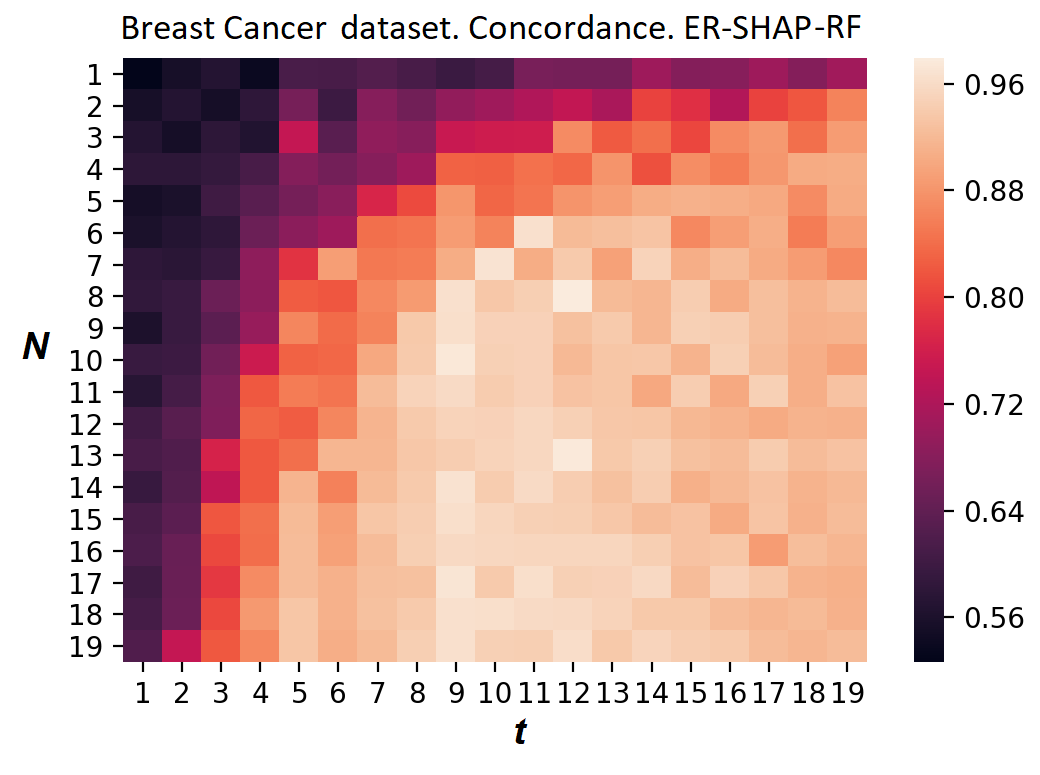}%
\caption{The heatmap reflecting the concordance index $C$ obtained by
ER-SHAP-RF for the Breast Cancer dataset with using the temperature scaling}%
\label{f:heatmap_breast_7}%
\end{center}
\end{figure}

Figs. \ref{f:heatmap_breast_6} and \ref{f:heatmap_breast_7} illustrate
heatmaps of the concordance index of ER-SHAP-RF for the Breast Cancer dataset.
They show results similar to the results obtained for the Boston Housing
dataset demonstrated in Figs. \ref{f:heatmap_boston_6} and
\ref{f:heatmap_boston_7}, respectively. This implies that the use of
\textquotedblleft pre-training\textquotedblright\ in the form of the random
forest combined with the calibration method leads to better approximation.%

%

\section{Conclusion}

It is important to note that only three modifications of the ensemble-based
SHAP have been presented. At the same time, many additional modifications of
the general approach based on constructing the ensemble of SHAPs can be
developed following the proposed modifications and the idea of the
ensemble-based approximation. 

First of all, the model of the feature selection used in ER-SHAP-RF for
\textquotedblleft pre-training\textquotedblright\ can be changed. There are
many methods solving the feature selection problem. Moreover, simple
explanation methods can be also applied to the preliminary selection of
important features and to computing their probability distribution. 

Second, various rules different from averaging can be applied to combining the
results of SHAPs, for example, the largest (smallest) Shapley values can be
computed for providing pessimistic (optimistic) decisions. 

The ensemble-based approach can be applied to explanation of the
classification as well as regression black-box models. It gives many
opportunities for developing new methods which can be viewed as directions for
further research. The proposed approach can be applied to local and global
explanations. However, its main advantage is that it significantly reduces the
computation time for solving the explanation problem. 

\section*{Acknowledgement}

The research is partially funded by the Ministry of Science and Higher
Education of the Russian Federation as part of World-class Research Center
program: Advanced Digital Technologies (contract No. 075-15-2020-934 dated 17.11.2020).

\bibliographystyle{plain}
\bibliography{Boosting,Classif_bib,Deep_Forest,Explain,Explain_med,FeatureSelection,Lasso,Math_bib,MYBIB,MYUSE,Robots,Survival_analysis}

\end{document}